\documentclass[acmtog,nonacm]{acmart}

\usepackage{fp}
\usepackage{expl3}[2012-07-08]
\ExplSyntaxOn
\ExplSyntaxOff

\usepackage{amsmath,amsfonts,bm}

\def\x{{x}}

\def\xi{{\x_i}}

\newcommand{\ignorethis}[1]{}

\def\eqref#1{equation~\ref{#1}}

\def\1{\bm{1}}

\def\vc{{\bm{c}}}

\def\vo{{\bm{o}}}

\def\vr{{\bm{r}}}

\def\vt{{\bm{t}}}

\def\vv{{\bm{v}}}

\def\vx{{\bm{x}}}

\DeclareMathAlphabet{\mathsfit}{\encodingdefault}{\sfdefault}{m}{sl}
\SetMathAlphabet{\mathsfit}{bold}{\encodingdefault}{\sfdefault}{bx}{n}

\newcommand{\ignore}[1]{}

\makeatletter
\DeclareRobustCommand\onedot{\futurelet\@let@token\@onedot}
\def\@onedot{\ifx\@let@token.\else.\null\fi\xspace}

\makeatother

\usepackage{multirow}
\usepackage{booktabs} % For formal tables
\usepackage{url}
\usepackage{amsmath,bm}
\usepackage[ruled]{algorithm2e} % For algorithms

\SetAlFnt{\small}
\SetAlCapFnt{\small}
\SetAlCapNameFnt{\small}
\SetAlCapHSkip{0pt}
\setcopyright{none}
\usepackage{overpic} % add text on image
\usepackage{subcaption}

\DeclareMathAlphabet{\mathcal}{OMS}{cmsy}{m}{n}

\usepackage{array}
\newcolumntype{L}[1]{>{\raggedright\let\newline\\\arraybackslash\hspace{0pt}}m{#1}}
\newcolumntype{C}[1]{>{\centering\let\newline\\\arraybackslash\hspace{0pt}}m{#1}}
\newcolumntype{R}[1]{>{\raggedleft\let\newline\\\arraybackslash\hspace{0pt}}m{#1}}

\newcommand{\Eref}[1]{Equation~(\ref{#1})}
\newcommand{\Fref}[1]{Figure~\ref{#1}}

\newcommand{\imgs}{\bm{I}_{src}}
\newcommand{\imgt}{\bm{I}_{tgt}}
\newcommand{\texts}{\vt_{src}}
\newcommand{\textt}{\vt_{tgt}}
\newcommand{\textn}{\vt_{neg}}
\newcommand{\Textn}{\mathcal{T}_{neg}}
\newcommand{\view}{\mathbf{d}}

\begin{document}
% Title portion
\title{\textit{NeRF-Art}: Text-Driven Neural Radiance Fields Stylization}

\author{Can Wang}
\affiliation{
  \institution{City University of Hong Kong}
}
\email{cwang355-c@my.cityu.edu.hk}
\author{Ruixiang Jiang}
\affiliation{
  \institution{The Hong Kong Polytechnic
University}
}
\email{cwang355-c@my.cityu.edu.hk}
\author{Menglei Chai}
\affiliation{
  \institution{Snap Inc.}
}
\email{cmlatsim@gmail.com}
\author{Mingming He}
\affiliation{
  \institution{Netflix}
}
\email{hmm.lillian@gmail.com}
\author{Dongdong Chen}
\affiliation{
  \institution{Microsoft Cloud AI}
}
\email{cddlyf@gmail.com}
\author{Jing Liao$^*$}\thanks{*Corresponding Author}
\affiliation{
  \institution{City University of Hong Kong}
}
\email{jingliao@cityu.edu.hk}

\begin{abstract}
As a powerful representation of 3D scenes, the neural radiance field (NeRF) enables high-quality novel view synthesis from multi-view images. Stylizing NeRF, however, remains challenging, especially on simulating a text-guided style with both the appearance and the geometry altered simultaneously. In this paper, we present \textit{NeRF-Art}, a text-guided NeRF stylization approach that manipulates the style of a pre-trained NeRF model with a simple text prompt. Unlike previous approaches that either lack sufficient geometry deformations and texture details or require meshes to guide the stylization, our method can shift a 3D scene to the target style characterized by desired geometry and appearance variations without any mesh guidance. This is achieved by introducing a novel global-local contrastive learning strategy, combined with the directional constraint to simultaneously control both the trajectory and the strength of the target style. Moreover, we adopt a weight regularization method to effectively suppress cloudy artifacts and geometry noises which arise easily when the density field is transformed during geometry stylization. Through extensive experiments on various styles, we demonstrate that our method is effective and robust regarding both single-view stylization quality and cross-view consistency. The code and more results can be found in our project page: \url{https://cassiepython.github.io/nerfart/}.
\end{abstract}



\begin{teaserfigure}
\centering
\setlength{\tabcolsep}{0\linewidth}
\includegraphics[width=\textwidth]{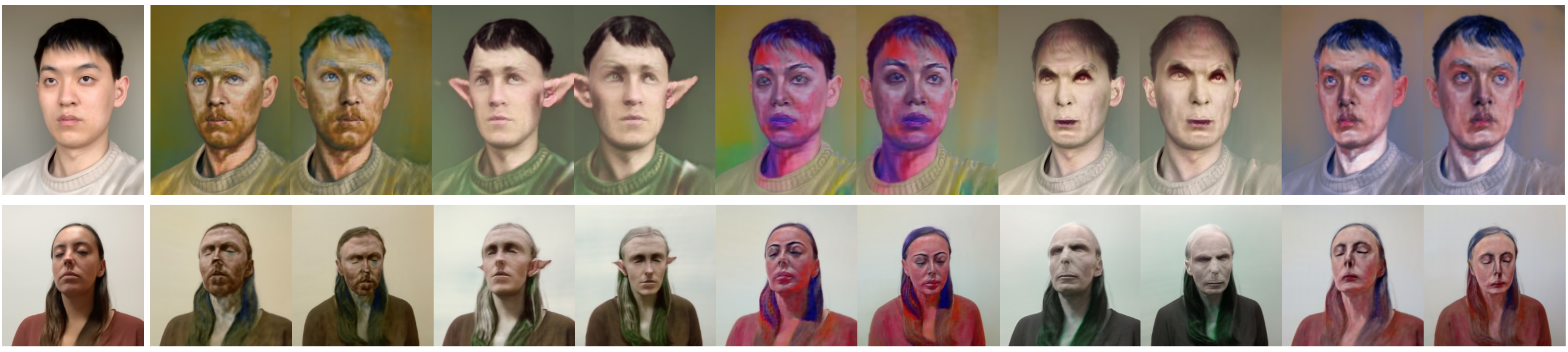}
\begin{small}
\begin{tabular}{C{0.09\linewidth}C{0.182\linewidth}C{0.182\linewidth}C{0.182\linewidth}C{0.182\linewidth}C{0.182\linewidth}}
Source&\textit{``Vincent van Gogh''}&\textit{``Tolkien Elf''}&\textit{``Fauvism''}&\textit{``Lord Voldemort''}&\textit{``Edvard Munch''}
\end{tabular}
\end{small}
\includegraphics[width=\textwidth]{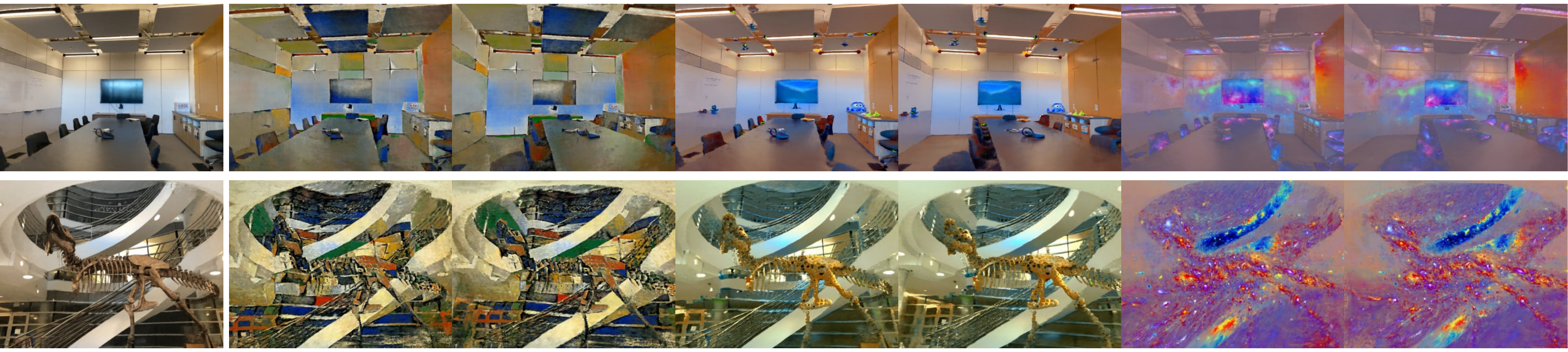}
\begin{small}
\begin{tabular}{C{0.13\linewidth}C{0.3\linewidth}C{0.3\linewidth}C{0.27\linewidth}}
Source&\textit{``Cubism Painting''}&\textit{``Pixar 3D Style''}&\textit{``Colorful Galaxy''}
\end{tabular}
\end{small}
\caption{\textbf{NeRF-Art Results.} Our \textit{NeRF-Art} stylizes a pre-trained NeRF to match the desired style described by a text prompt. It modulates not only appearance but also geometry of NeRF.}
\label{fig:teaser}
\end{teaserfigure}


\maketitle

\section{Introduction}

Artistic works depict the world in various creative and imaginative styles, evolving along with human progress. While primarily driven by professionals, the generation of artistic content is now more accessible to average users than ever before, empowered by the recent research on visual artistic stylization. In the era of deep learning, technical advances are gradually reshaping how people create, consume, and share art, from real-time entertainment to concept design. Since neural style transfer~\cite{gatys2016image,chen2017stylebank,shu2021gan,zhao2014parallel,sheng2018deep} shows the potential of encoding and changing visual styles via deep neural networks, a significant amount of effort has been devoted to effectively and efficiently migrating the style of an arbitrary image~\cite{gatys2016image,huang2017arbitrary,li2017universal,liao2017visual} or a specific domain~\cite{zhu2017unpaired,lee2020drit++} to the content image. Despite impressive results, these methods are limited to stylizing a single view captured by the content image.

Motivated by the increasing demand for 3D asset creation, our goal is to stylize \textit{3D content} from \textit{multi-view input}, in contrast to single-image stylization. In the domain of 3D representation, previous methods typically take explicit models (e.g., meshes~\cite{kato2018neural,hollein2021stylemesh,han2021exemplarbased,ye20213d,zhang2020deep}, voxels~\cite{guo2021volumetric,klehm2014property}, and point clouds~\cite{cao2020psnet,lin2018learning}) followed by differentiable rendering for multi-view stylization. These methods enable intuitive control over the geometry but suffer from the limited capacity for modeling and rendering complex scenes. The recent implicit representation of neural radiance field~(NeRF)~\cite{mildenhall2020nerf,deng2022fov,yang2022recursive,zhang2022controllable,wang2022nerfcap} significantly improves the quality of novel view synthesis, satisfying our needs for a general representation of various scenes and objects. However, while enjoying the superior scene reconstruction quality of NeRF, the curse of its highly implicit volumetric representation of appearance and geometry, parameterized and entangled by dense MLP networks, makes NeRF more challenging to stylize through jointly transforming the encoded color and shape.

Very recently, pioneering NeRF stylization works~\cite{chiang2022stylizing,fan2022unified} have made exhilarating progress on appearance style transfer of 3D scenes. However, their style guidance is limited to image reference, which, although being adopted as one common way to specify the target style, is not always a perfect solution for every scenario---obtaining appropriate style images that both reflect the target style and match the source content might not be easy or even possible in many cases. Therefore, finding another simple, natural, and expressive form of guidance becomes an attractive idea. Thanks to the parallel advances in language-vision models, stylization with natural language is no longer a fantasy. As demonstrated by recent text-guided stylization works~\cite{gal2021stylegan,wei2021hairclip,michel2021text2mesh,Hong2022Avatar}, compared to image-guided approaches, short text prompts provide 1) an extremely intuitive and user-friendly way to specify styles, 2) a flexible control over various styles from abstract ones like a certain concept to very concrete ones like a famous painting or character, and 3) a view-independent representation that is free 
from content alignment and naturally benefits cross-view consistency.

\begin{figure*}[ht!]
\centering
\includegraphics[width=0.9\linewidth]{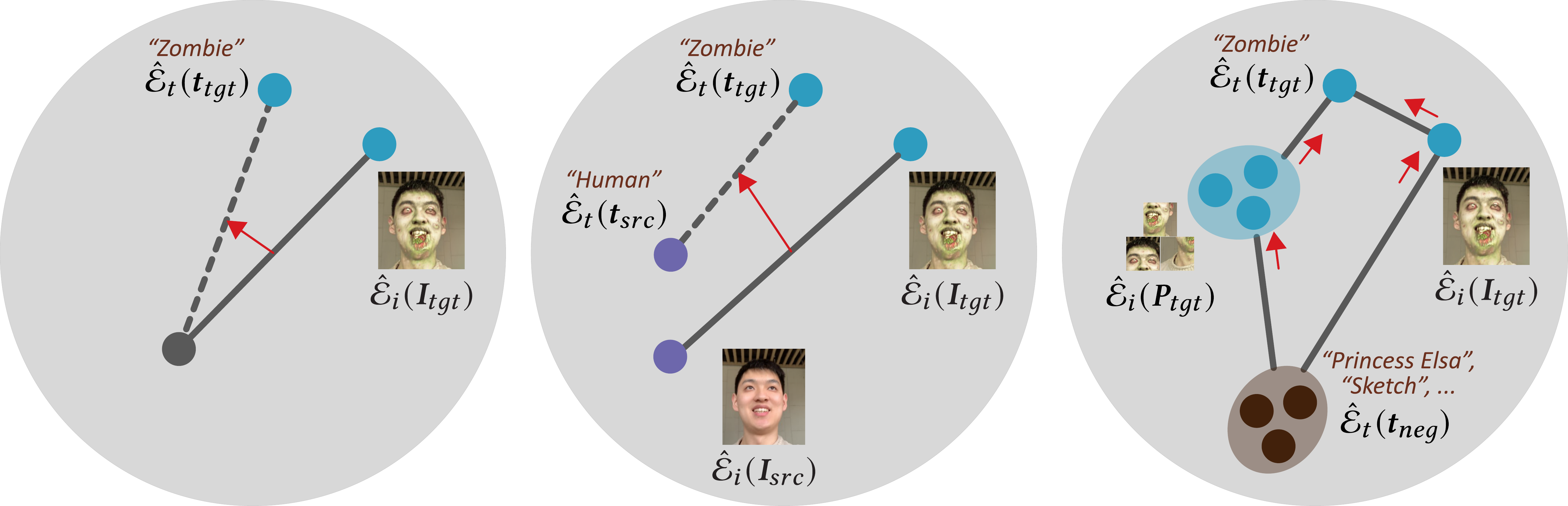}
\setlength{\tabcolsep}{0\linewidth}
\begin{tabular}{C{0.26\linewidth}C{0.28\linewidth}C{0.26\linewidth}}
(a) $\mathcal{L}_{dir}^a$&(b) $\mathcal{L}_{dir}^r$&(c) $\mathcal{L}_{con}^l$ \& $\mathcal{L}_{con}^g$\\
\end{tabular}
\caption{\textbf{CLIP-Guided Stylization Losses.} (a) The absolute directional loss; (b) The relative directional loss; (c) The global and local contrastive loss.}
\label{fig:loss}
\end{figure*}

Yet, with the existing approaches, it is still challenging to stylize the implicit representation of NeRF via a simple text prompt.  
Learning a latent space helps constrain the geometry and texture modulations~\cite{wang2021clip}, but it is often data-dependent and laborious. Some efforts directly enforce style directions (\Fref{fig:loss}) between the rendered views of NeRF and the text in the CLIP~\cite{radford2021learning} embedding space. In addition, background augmentation~\cite{jain2022zero} and mesh guidance~\cite{Hong2022Avatar} have been proposed to improve the geometry and texture modulations. 
However, they still suffer from insufficient geometry deformations and texture details. 

In this work, we propose \textit{NeRF-Art}, a new text-driven NeRF stylization method.
Given a pre-trained NeRF model and a single text prompt, our method enables consistent novel view synthesis with both appearance and geometry transformed, adhering to the specified style. This is achieved by combining the recent large-scale Language-Vision model (i.e., CLIP) with NeRF, which is non-trivial due to several challenges. Directly applying the supervision from CLIP to NeRF by constraining the similarity between the rendered views and the text in the embedding space as~\cite{gal2021stylegan} is insufficient to ensure the desired style strength. To tackle this problem, we design a CLIP-based contrastive loss to properly strengthen the stylization, by bringing the results closer to the target style and farther away from other styles pre-defined as negative samples. To further ensure the uniformity of the style over the whole scene, we extend our contrastive constraint to a hybrid global-local framework to cover both global structures and local details. In addition, to support geometry stylization jointly with appearance, we relax the constraints on the density of the pre-trained NeRF and adopt a weight regularization to effectively reduce cloudy artifacts and geometry noises when altering the density field.
In experiments, we first evaluate text description selection for stylization and then test our method on various styles and demonstrate text guidance's effectiveness and flexibility for NeRF stylization.
Furthermore, we conduct a user study to show that our method achieves the best visual-pleasing results compared to related methods.
We also extract the mesh from the stylized NeRF to show the geometry modulation ability of our method and integrate with different baselines to demonstrate the generalization ability of our method to various NeRF-like models.

\section{Related Work}

\noindent{\textbf{Neural Style Transfer on Images and Videos.}}  
Artistic image stylization is a long-standing research area. Traditional methods use handcrafted features to simulate styles~\cite{hertzmann1998painterly,hertzmann2001image}. With the fast development of deep learning, neural networks have been applied to style transfer from either an arbitrary image~\cite{gatys2016image,johnson2016perceptual,huang2017arbitrary,liao2017visual,li2017universal,kolkin2019style} or a specific domain~\cite{zhu2017unpaired,huang2018multimodal,huang2021unsupervised,lee2020drit++}, and achieved impressive results. By enforcing temporal smoothness constraints defined on optical flows, neural style transfer has been successfully extended to videos~\cite{ruder2016artistic,chen2017coherent,chen2020optical}. However, both image and video stylization methods are restricted to the given views. Simply combining the neural style transfer and novel view synthesis methods without considering 3D geometry will lead to blurriness or view inconsistencies.

\begin{figure*}[ht!]
\includegraphics[width=0.8\linewidth]{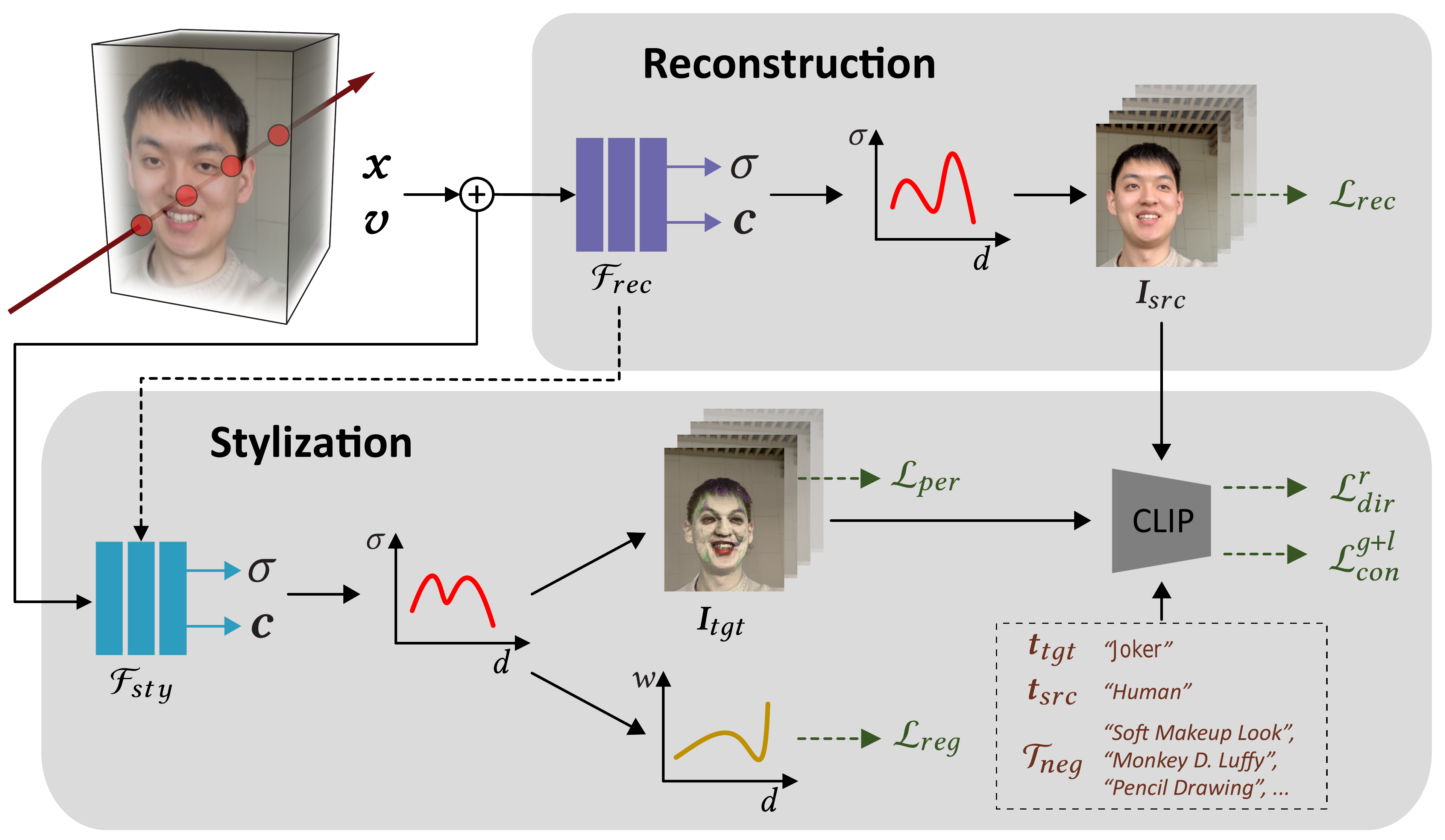}
\caption{\textbf{NeRF-Art Pipeline.} In the \textit{reconstruction} stage, our method first pre-trains the NeRF model $\mathcal{F}_{rec}$ of the target scene from multi-view input with reconstruction loss $\mathcal{L}_{rec}$. In the \textit{stylization} stage, our method stylized NeRF model $\mathcal{F}_{rec}$ to $\mathcal{F}_{sty}$, guided by a text prompt $\textt$, using a combination of relative directional loss $\mathcal{L}_{dir}^r$ and global-local contrastive loss $\mathcal{L}_{con}^{g+l}$ in the CLIP embedding space, plus weight regularization loss $\mathcal{L}_{reg}$ and perceptual loss $\mathcal{L}_{per}$.}
\label{fig:framework}
\end{figure*}

\noindent{\textbf{Neural Stylization on Explicit 3D Representations.}}
With the increasing demand for 3D content, neural style transfer has been extended to explicit 3D representations. The work \cite{chen2018stereoscopic} first considers the cross-view disparity consistency and applies style transfer on stereoscopic images or videos. Later, considering the voxel is the most compatible representation for CNNs, \textit{SKPN}~\cite{guo2021volumetric} encodes volume using convolutional blocks and stylizes it by deep features extracted from a reference image. As for mesh stylization, differential rendering allows for backpropagating style transfer objectives from rendered images to 3D meshes. According to whether the geometry or texture are allowed to be optimized, existing mesh style transfer methods achieve three different effects: texture stylization~\cite{mordvintsev2018differentiable,hollein2021stylemesh}, geometric stylization~\cite{liu2018paparazzi}, and joint stylization~\cite{kato2018neural,han2021exemplar,yin20213dstylenet}. Another line of work uses point clouds as the 3D proxy to guarantee 3D consistency in stylizing novel views from either a single image~\cite{mu20213d} or multiple frames~\cite{huang2021learning}. In these works, point-wise features extracted from pre-trained PointNet~\cite{qi2017pointnet} or GCN~\cite{li2021deepgcns} are stylized by feature transform algorithms, e.g., adaptive normalization, and then rendered to novel views. Despite the successes, these 3D stylization methods are difficult to generalize to complicated objects or scenes with dedicated structures, limited by the expressiveness of explicit 3D representations.

\noindent{\textbf{Neural Stylization on NeRF.}} 
To address the inherent limitations of explicit representations, implicit methods have recently received much attention. NeRF is a seminal one that is able to represent complex scenes by parameterizing the implicit function as MLP networks. A large number of follow-up works are presented to improve its efficiency~\cite{deng2021depth,lindell2021autoint,garbin2021fastnerf,reiser2021kilonerf,yu2021plenoctrees,muller2022instant}, quality~\cite{barron2021mip,arandjelovic2021nerf,ma2021deblur,zhang2020nerf++}, controllablity~\cite{zhang2021nerfactor,srinivasan2021nerv,liu2021editing,wang2021clip}, and generalization~\cite{jain2021putting,yu2021pixelnerf,niemeyer2021regnerf,park2021nerfies,pumarola2021d,park2021hypernerf,tretschk2021non,noguchi2021neural,peng2021animatable,li2021neural,xian2021space,gao2021dynamic}. Inspired by the power of NeRF, three very recent works~\cite{chiang2022stylizing,Huang22StylizedNeRF,zhang2022arf} adopt it for 3D stylization. They design the stylization network to predict color-related parameters in the NeRF model based on a reference style. And the stylization network is trained either by imposing the image style transfer losses~\cite{gatys2016image,zhang2022arf} on rendered views~\cite{chiang2022stylizing} or being supervised by a mutually learnt image stylization network~\cite{Huang22StylizedNeRF}. These works have achieved consistent results in novel-view stylization. However, their stylization is still restricted to appearance only because they do not adjust density parameters in the NeRF model. In contrast, our method supports both appearance and geometric stylization to better mimic the reference style.
Moreover, they rely on reference images for stylization, while we seek to stylize the scenes via simple text prompts.

\noindent{\textbf{Text-Driven Stylization.}} 
Compared to image references, a natural language prompt is a more intuitive and user-friendly way to specify the style. Therefore, a current line of works shifted away from image reference towards text guidance, with the help of the pre-trained CLIP~\cite{radford2021learning}, which bridges texts and images by jointly learning a shared latent space. The pioneering work \textit{StyleGAN-NADA}~\cite{gal2021stylegan} proposes a directional CLIP loss for transferring the pre-trained StyleGAN2 model~\cite{karras2020analyzing} to the target domain with the desired style described by a textual prompt. However, it is an image-based method and will lead to inconsistencies when applied to stylizing multiple views. In the 3D world, \textit{Text2Mesh}~\cite{michel2021text2mesh} uses CLIP to guide the stylization of a given 3D mesh by learning a displacement map for geometry deformation and vertex colors for texture stylization. The contemporary work \textit{AvatarCLIP}~\cite{Hong2022Avatar} further supports driving a stylized human mesh using natural languages. Despite their success, 
these methods are limited to mesh input. In contrast, our method is able to stylize 3D scenes with better visual quality and view consistency without any mesh input.

\section{Overview}

As illustrated in \Fref{fig:framework}, our approach is simply decomposed into reconstruction and stylization stages.
In what follows, after briefly reviewing our 3D photography representation with NeRF (\S~\ref{subsec:Preliminary}),
we focus on introducing our text-guided stylization method.
Specifically, we first formulate the directional CLIP loss for stylization, which leverages the power of the pre-trained Language-Vision model (\S~\ref{subsec:Stylization}). Then, we introduce our global-local contrastive learning framework to cope with the stylization strength issue of the directional CLIP loss (\S~\ref{subsec:Contrastive}). Next, we introduce a weight regularization term to alleviate the cloudy artifacts caused and geometry noises by the stylization process (\S~\ref{subsec:weight}). Finally, we conclude this section with the overall training strategy of the entire pipeline (\S~\ref{subsec:Training}).

\subsection{Preliminary on NeRF Scene Representation}
\label{subsec:Preliminary}
We take NeRF as our 3D scene representation, which defines a continuous volumetric field as implicit functions, parameterized by MLP networks $\mathcal{F}$.
Given a single spatial coordinate $\vx=(x,y,z)$ and its corresponding view direction $\view=(\phi,\theta)$, the network predicts the density $\sigma$ and view-dependent radiance $\vc=(r,g,b)$, leading to the final color $C(\bm{r})$ of the camera ray $\bm{r}(t)=\vo+t\view$ by accumulating $K$ sample points along it, given the target view:
\begin{equation}
C(\bm{r})=\sum\nolimits_{k=1}^{K}T_k(1-\omega_k)\vc_k,
\label{eq:nerf_render}
\end{equation}
where $\omega_k=\exp(-\sigma_k(d_{k+1}-d_k))$ represents the transmittance of the ray segment $(k,k+1)$ and $\begin{matrix}T_k=\prod_i^{k-1}\omega_i\end{matrix}$ is the accumulated transmittance from the origin to the sample $k$.

To train NeRF from a set of multi-view photos, a simple supervised reconstruction loss is adopted between the ground-truth pixel colors $\hat{C}(\bm{r})$ from the training view and the NeRF prediction $C(\bm{r})$:
\begin{equation}
\mathcal{L}_{rec}=\sum\nolimits_{\bm{r}}\big\|C(\bm{r})-\hat{C}(\bm{r})\big\|_2^2.
\label{eq:nerf_mse}
\end{equation}

\section{Text-Guided NeRF Stylization}

After optimizing the reconstructed NeRF model $\mathcal{F}_{rec}$ from the multi-view input (\S~\ref{subsec:Preliminary}), our goal is to train a stylized NeRF model $\mathcal{F}_{sty}$, which satisfies the style control of the target text prompt $\textt$ while preserving the content from $\mathcal{F}_{rec}$ (\Fref{fig:framework}).

The CLIP model aligns the semantics of image and text in a joint embedding space, by utilizing the image encoder $\hat{\mathcal{E}}_i(\cdot)$ and the text encoder $\hat{\mathcal{E}}_t(\cdot)$. The semantic power of CLIP bridges the gap between natural language prompts and synthesized image pixels, making it possible to stylize NeRF scenes with text controls.

However, even with the powerful embedding space of CLIP, it remains challenging to achieve text-guided NeRF stylization that 1) preserves the original content from being washed away by the new style, 2) reaches the target style with proper strength that satisfies the semantics of the input text prompt, and 3) maintains cross-view consistency and avoids artifacts in the final NeRF model.

\subsection{\textit{Trajectory} Control w/ Directional CLIP Loss}
\label{subsec:Stylization}

An intuitive strategy for text-guided NeRF stylization would be to enforce the trajectory of the stylization in the CLIP space with an \textit{absolute} directional CLIP loss that measures the cosine similarity ($\langle\cdot,\cdot\rangle$) between the stylized NeRF rendering $\imgt$ and the target text prompt $\textt$ (\Fref{fig:loss}(a)):
\begin{equation}
\mathcal{L}_{dir}^a=\sum\nolimits_{\imgt}\Big[1-\big\langle\hat{\mathcal{E}}_i(\imgt),\hat{\mathcal{E}}_t(\textt)\big\rangle\Big],
\label{eq:clip_global}
\end{equation}
which guides NeRF rendering with a global direction of the target text, not depending on any reference starting point. This loss is first designed in \textit{StyleCLIP}~\cite{patashnik2021styleclip} to guide face image editing and further extended to generative NeRF editing in \textit{CLIP-NeRF}~\cite{wang2021clip}.

However, as observed in \textit{StyleGAN-NADA}~\cite{gal2021stylegan}, this global loss could easily mode-collapse the generator and hurt the generation diversity of stylization.
Therefore, a \textit{relative} directional loss is proposed, which transfers the source image $\imgs$ to the target domain guided by the CLIP-space trajectory embedded by a pair of text prompts $(\texts,\textt)$ instead of a single one (\Fref{fig:loss}(b)).
This relative directional CLIP loss for our NeRF stylization is defined as:
\begin{equation}
\mathcal{L}_{dir}^r=\sum\nolimits_{\imgt}\Big[1-\big\langle\hat{\mathcal{E}}_i(\imgt)-\hat{\mathcal{E}}_i(\imgs),\hat{\mathcal{E}}_t(\textt)-\hat{\mathcal{E}}_t(\texts)\big\rangle\Big].
\label{eq:clip_direction}
\end{equation}
Different from the single-image setting of \textit{StyleGAN-NADA}, here, the training target $\imgt$ stands for an arbitrarily sampled view rendered by the stylized NeRF of the same scene, and the source image $\imgs$ is produced by the pre-trained NeRF model and shares the identical view as $\imgt$. We will follow this convention hereinafter.

\subsection{\textit{Strength} Control w/ Glocal Contrastive Learning}
\label{subsec:Contrastive}

As the directional CLIP loss (\Eref{eq:clip_direction}) works by measuring the similarity between the normalized unit directions of the embedded vectors, it can enforce the relative stylization trajectory. However, it struggles with preserving enough stylization strength in altering the pre-trained NeRF model.

To address this issue, we propose a contrastive learning strategy to control the stylization strength (\Fref{fig:loss}(c)). Specifically, in the framework of contrastive learning, with the rendered view $\imgt$ as the query target, we set positive samples to the target text prompt $\textt$ with the desired style and construct negative samples $\textn\in\Textn$ by sampling a set of text prompts semantically irrelevant to $\imgt$.
In general, our contrastive loss in the CLIP space is defined as:
\begin{equation}
\mathcal{L}_{con}=-\sum\nolimits_{\imgt}\log\Bigg[\frac{\exp(\vv\cdot\vv^+/\tau)}{\exp(\vv\cdot\vv^+/\tau)+\sum_{\vv^-}\exp(\vv\cdot\vv^-/\tau)}\Bigg],
\label{eq:contrast} 
\end{equation}
where $\{\vv,\vv^+,\vv^-\}$ are query, positive sample, and negative sample, respectively, and temperature $\tau$ is set to $0.07$ in all our experiments. When defining the loss globally by treating the entire view $\imgt$ as the query anchor, we have the global contrastive loss $\mathcal{L}_{con}^g$ with $\{\vv=\hat{\mathcal{E}}_i(\imgt),\,\vv^+=\hat{\mathcal{E}}_t(\textt),\,\vv^-=\hat{\mathcal{E}}_t(\vt_{neg})\}$.

Ideally, this global contrastive loss cooperates with the directional CLIP loss, where the former defines the style trajectory that aligns with the target text, and the latter, at the same time, ensures the proper stylization magnitude by pushing along the style trajectory.
However, the global contrastive loss still has trouble achieving sufficient and uniform stylization on the entire NeRF scene, leading to excessive stylization on certain parts and insufficient stylization in other regions.
This may be attributed to the fact that CLIP focuses more attention on local regions with distinguishable features than the entire scene. Thus, this global contrastive loss can deliver a small value even when the overall stylization is insufficient or non-uniform.
To achieve a more sufficient and balanced stylization, enforced by a more locally-attended contrastive learning approach, inspired by \textit{PatchNCE} loss~\cite{park2020contrastive}, we propose a complementary local contrastive loss $\mathcal{L}_{con}^l$ which sets queries to random local patches $\bm{P}_{tgt}$ cropped from $\imgt$: $\{\vv=\hat{\mathcal{E}}_i(\bm{P}_{tgt}),\,\vv^+=\hat{\mathcal{E}}_t(\textt),\,\vv^-=\hat{\mathcal{E}}_t(\vt_{neg})\}$.

Overall, we combine the global and local terms as our final global-local contrastive loss:
\begin{equation}
\mathcal{L}_{con}^{g+l}=\lambda_g\mathcal{L}_{con}^g+\lambda_l\mathcal{L}_{con}^l.
\label{eq:contrast}
\end{equation}

\subsection{Artifact Suppression w/ Weight Regularization}
\label{subsec:weight}

Our pipeline aims to change not only the color but also the density of the pre-trained NeRF to achieve a joint stylization of appearance and geometry. However, allowing the training process to alter the density may lead to cloud-like semi-transparent artifacts near the camera and geometry noises, even if the pre-trained NeRF is perfectly clean. To alleviate that, we adopt a weight regularization loss to suppress geometric noises and encourage a more concentrated density distribution that better resembles real-world scenes.

Based on our NeRF notations (\Eref{eq:nerf_render}), weight of each ray sample is defined as the contribution to the final ray color: $w_k=T_k(1-\omega_k)$, where $\sum_k w_k\leq 1$. Similar to the distortion loss in \textit{mip-NeRF 360}~\cite{barron2022mipnerf360}, the weight regularization loss is defined as:
\begin{equation}
\mathcal{L}_{reg}=\sum\nolimits_{\imgt}\sum\nolimits_{\vr}\sum\nolimits_{(i,j)\in K}w_iw_j\left\|d_i-d_j\right\|,
\end{equation}
where for each ray $\vr$ of a randomly sampled view $\imgt$, pairs of samples $(i,j)$ with distances $\|d_i-d_j\|$ are sampled.
But different from \textit{mip-NeRF 360}~\cite{barron2022mipnerf360} that optimizes the distances, we penalize those pairs with scattered large weights to suppress noise peeks and aggregate weights to the correct object surface.

\subsection{Training}
\label{subsec:Training}

During training, we finetune the pre-trained NeRF model for stylization. The overall objective consists of three parts: text-guided stylization losses (including directional CLIP loss and global-local contrastive loss to control style trajectory and strength, respectively), content-preservation loss (we adopt VGG-based perceptual loss), and artifact suppression regularization loss:
\begin{equation}
\mathcal{L}=(\mathcal{L}_{dir}^r+\mathcal{L}_{con}^{g+l})+\lambda_p\mathcal{L}_{per}+\lambda_r\mathcal{L}_{reg}.
\label{eq:stylization}
\end{equation}
Here we define the perceptual loss $\mathcal{L}_{per}$ between the original and stylized NeRF renderings on certain pre-defined VGG layers $\psi\in\Psi$:
\begin{equation}
\mathcal{L}_{per}=\sum\nolimits_{\imgt}\sum\nolimits_{\psi\in\Psi}\|\psi(\imgt)-\psi(\imgs)\|^2_2.
\label{eq:perceptual}
\end{equation}

It's practically infeasible to train stylization on all rays due to backward gradient propagation's prohibitively huge memory consumption. To address this issue, previous works either sample sparse rays to obtain coarse images or patches~\cite{schwarz2020graf,chiang2022stylizing,jain2021putting,Hong2022Avatar} or render all rays to low resolution and then upsample with CNN networks~\cite{niemeyer2021giraffe}. However, coarse renderings or patches lose style details and semantic structures, while upsampling harms the cross-view consistency.
Instead, we adopt a much easier solution, which first renders all rays to obtain the whole image of an arbitrary view, calculates the stylization loss gradients in the forward process, and then back-propagates the gradients through NeRF at the patch level. This significantly reduces memory consumption and allows rendering high-resolution images for better stylization training. 

\begin{figure}[t]
\setlength{\tabcolsep}{0\linewidth}
\centering
\includegraphics[width=0.88\linewidth]{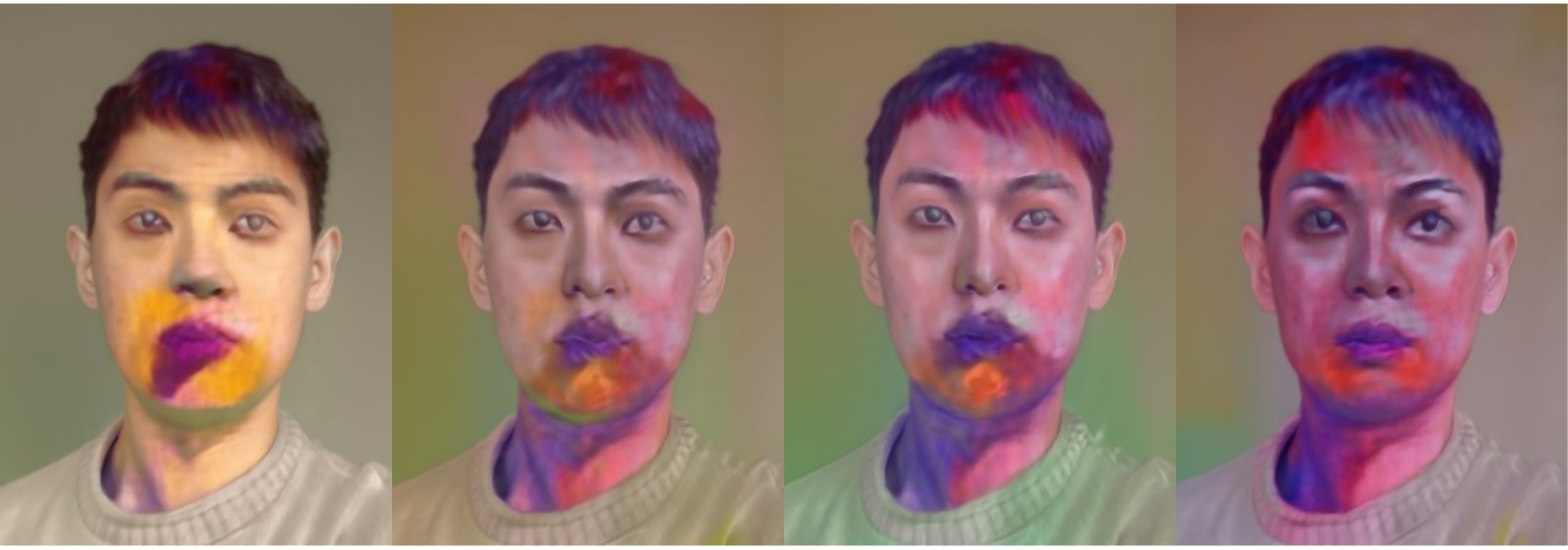}
\begin{scriptsize}
\begin{tabular}{C{0.22\linewidth}C{0.22\linewidth}C{0.22\linewidth}C{0.22\linewidth}}
\textit{``Fauvism''}&\textit{``Fauvism painting''}&\textit{``painting, Fauvism style''}&\textit{``painting, oil on canvas, Fauvism style''}
\end{tabular}
\end{scriptsize}
\includegraphics[width=0.88\linewidth]{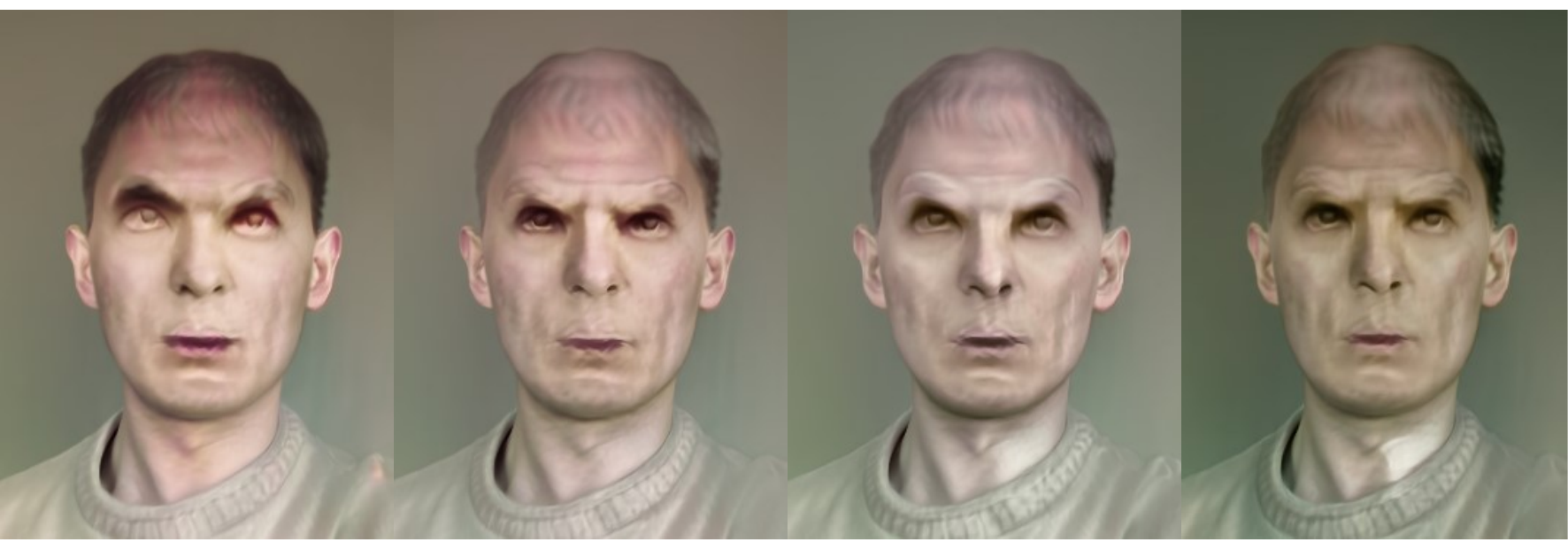}
\begin{scriptsize}
\begin{tabular}{C{0.22\linewidth}C{0.22\linewidth}C{0.22\linewidth}C{0.22\linewidth}}
\textit{``Lord Voldemort''}&\textit{``Head of Lord Voldemort''}&\textit{``Head of Lord Voldemort in fantasy style''}&\textit{``Head of Lord Voldemort in fantasy style, Harry Potter Film''}
\end{tabular}
\end{scriptsize}
\includegraphics[width=0.88\linewidth]{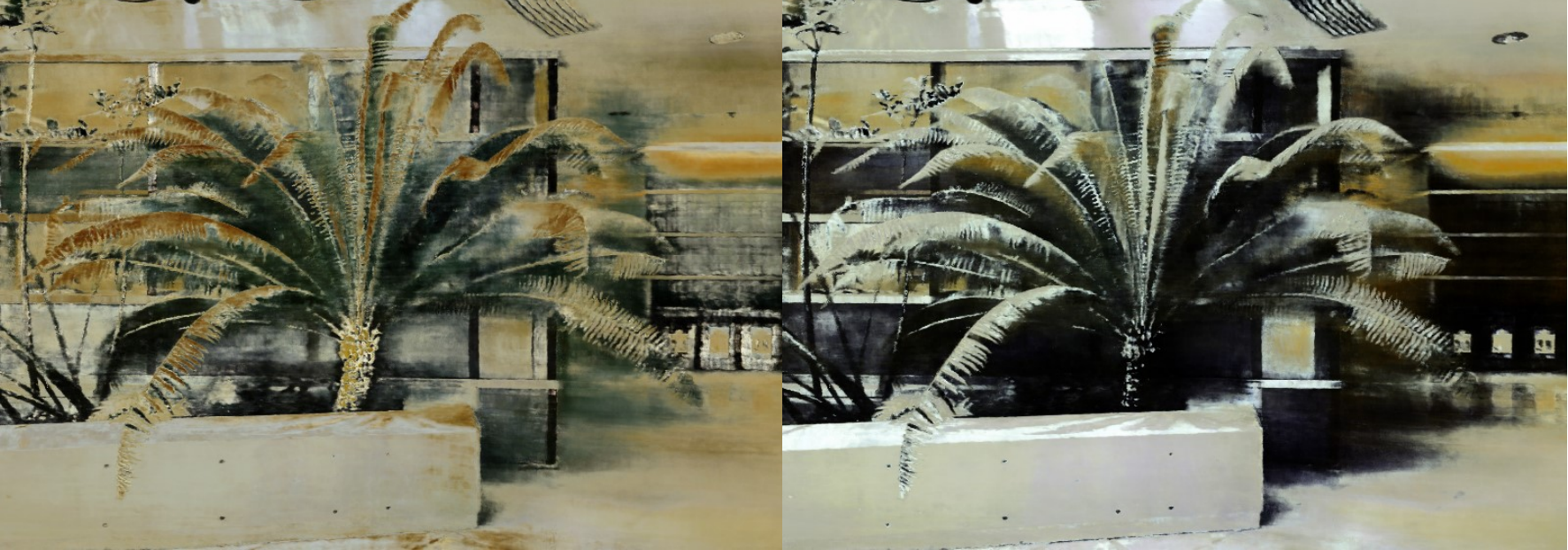}
\begin{scriptsize}
\begin{tabular}{C{0.44\linewidth}C{0.44\linewidth}}
\textit{``Chinese Painting''}&\textit{``Chinese Ink Painting''}
\end{tabular}
\end{scriptsize}
\includegraphics[width=0.88\linewidth]{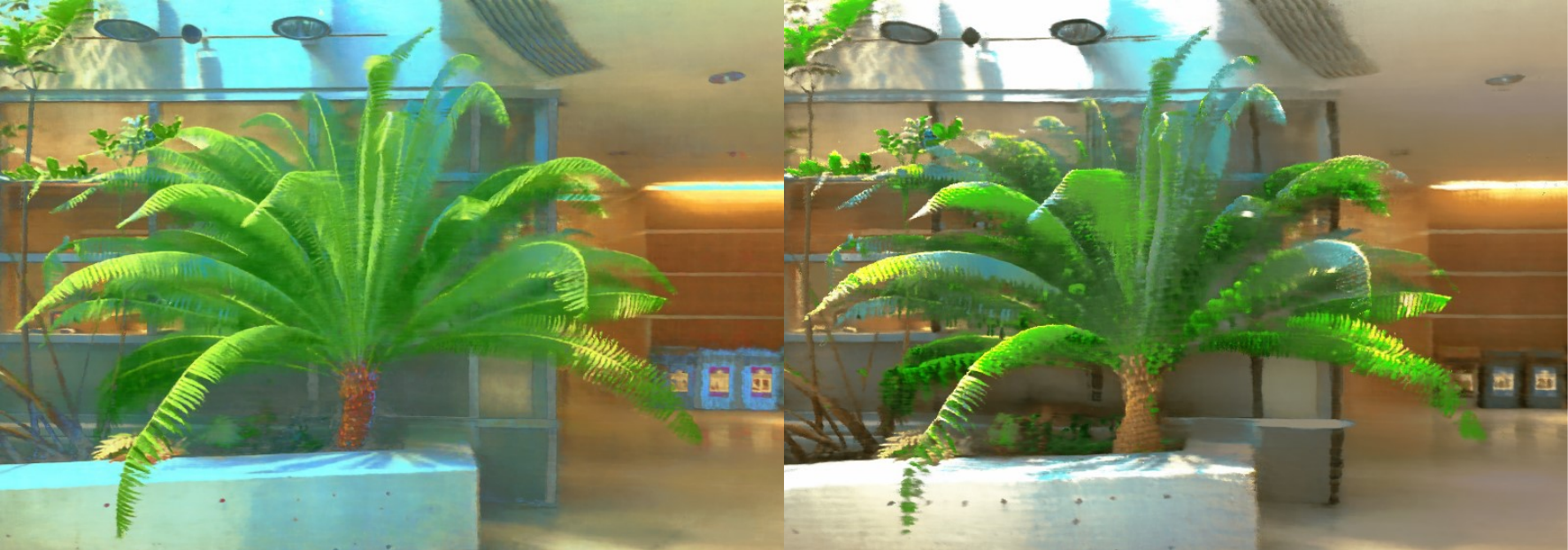}
\begin{scriptsize}
\begin{tabular}{C{0.44\linewidth}C{0.44\linewidth}}
\textit{``Pixar Style''}&\textit{``3D Render in the Style of Pixar''}
\end{tabular}
\end{scriptsize}
\caption{\textbf{Text Evaluation.} We present descriptions at different detail levels for a specific style. }
\label{fig:te}
\end{figure}

\begin{figure}[t]
\centering
\setlength{\tabcolsep}{0\linewidth}
\begin{small}
\begin{tabular}{C{0.20\linewidth}C{0.40\linewidth}C{0.40\linewidth}}
Source&\textit{``Joker''}&\textit{``Fernando Botero''}
\end{tabular}
\includegraphics[width=0.49\textwidth]{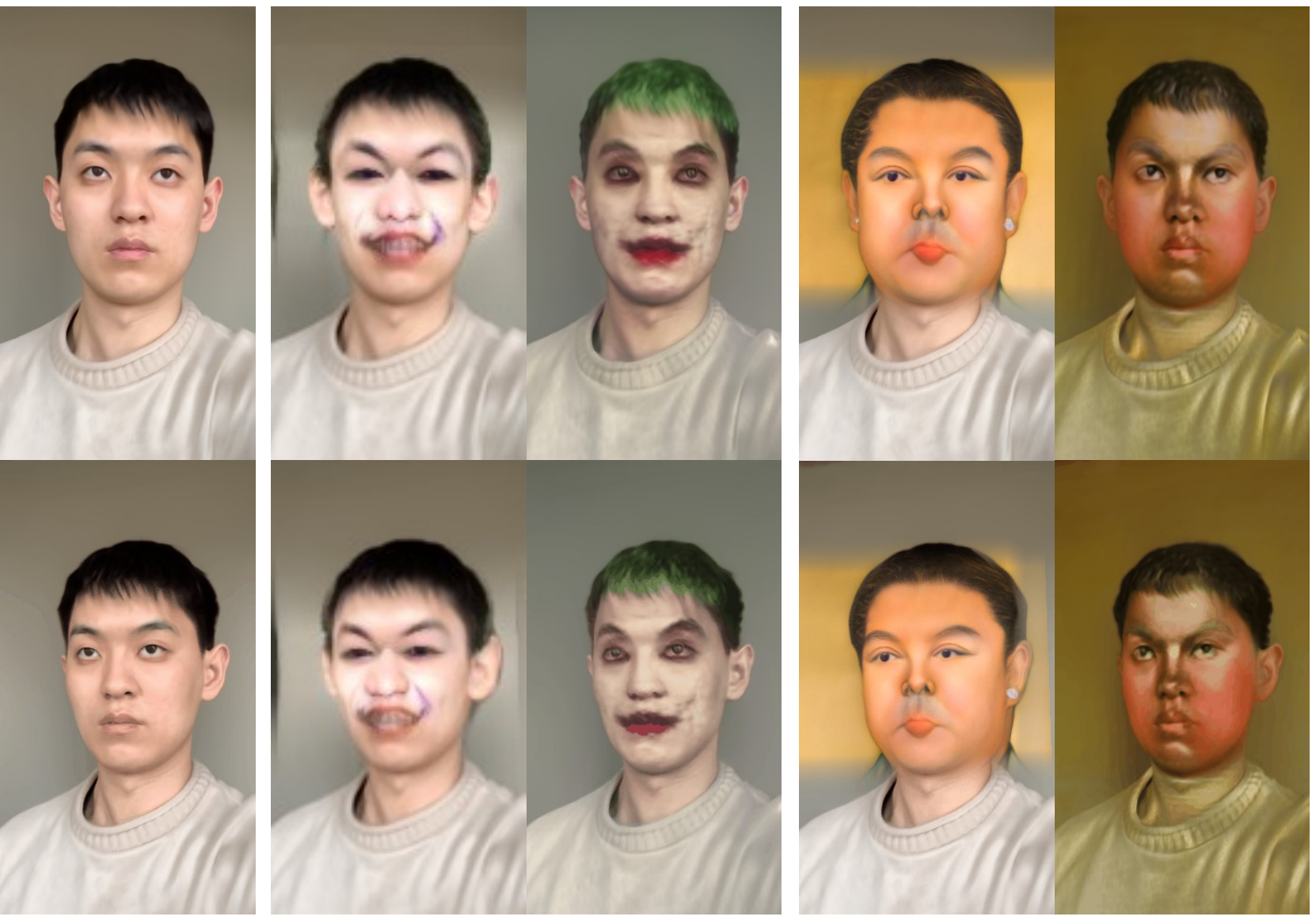}
\begin{tabular}{C{0.20\linewidth}C{0.40\linewidth}C{0.40\linewidth}}
Source&\textit{``Bear''}&\textit{``Vincent van Gogh''}
\end{tabular}
\includegraphics[width=1.0\linewidth]{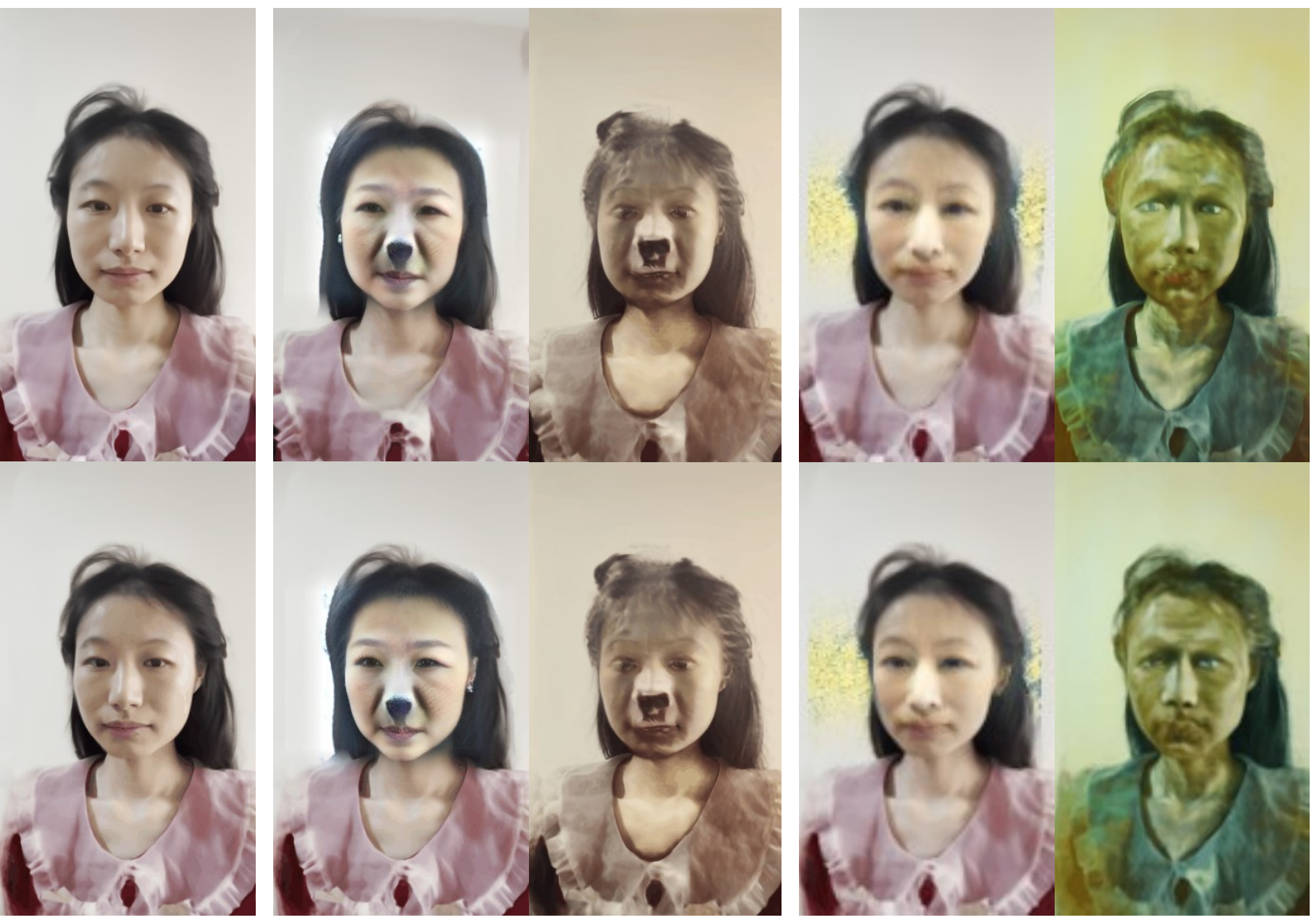}
\begin{tabular}{C{0.22\linewidth}C{0.18\linewidth}C{0.2\linewidth}C{0.22\linewidth}C{0.18\linewidth}}
&StyleGAN-NADA&Ours&StyleGAN-NADA&Ours
\end{tabular}
\end{small}
\caption{\textbf{Comparisons.} Comparisons with the text-guided image stylization method \textit{StyleGAN-NADA}~\cite{gal2021stylegan}.}
\label{fig:cmp1}
\end{figure}

\section{Experiments}
\label{sec:Exps}

\subsection{Implementation Details} We implement our framework using \textit{Pytorch}. In the reconstruction training stage, we sample $192$ points for each ray and train our model for $6$ epochs.
We set the learning rate as $0.0005$ and adopt the Adam optimizer. 
While in the stylization training stage, we train our model for $4$ epochs with the learning rate of $0.001$ and use the \textit{Adam} optimizer. We set hyper-parameters $\lambda_g$, $\lambda_l$, $\lambda_p$ and $\lambda_r$ as $0.2$, $0.1$, $2.0$, and $0.1$, respectively.
To construct the negative samples, we manually collect around 200 text descriptions from \textit{Pinterest} website, describing various styles, like \textit{``Zombie''}, \textit{``Tolkien elf''}, and \textit{``Self-Portrait by Van Gogh''}. 
We set the patch size as the $1/10$ of the original input in the local contrastive loss.
Without loss of generality, we adopt VolSDF~\cite{yariv2021volume} as the basic NeRF model for stylization.

\subsection{Data Collection}

Three self-portrait datasets are gathered under an in-the-wild condition by asking three users to capture selfies video for around 10 seconds with the front-facing camera.
We finally received six video clips in around 10 seconds. After collecting these video clips under different views and expressions, we extract 100 frames for each video clip using FFmpeg with 15 fps. Then these frames are resized to 270$\times$ 480. Then we estimate camera poses for these frames using COLMAP~\cite{schonberger2016structure} with rigid relative camera pose constraints. We suppose frames in a video share the same intrinsics. We also reconstruct a lady from the H3DS dataset~\cite{ramon2021h3d}. We remove noise frames and obtain 31 sparse views. Moreover, we use the image size with 256$\times$256 for stylization.
We also adopt the Local Light Field Fusion (LLFF) dataset~\cite{mildenhall2019local} to stylize non-face scenes. LLFF dataset is composed of forward-facing scenes, with around 20 to 60 images. 

\begin{figure*}[t]
\centering
\setlength{\tabcolsep}{0\linewidth}
\begin{small}
\begin{tabular}{C{0.11\linewidth}C{0.18\linewidth}C{0.2\linewidth}C{0.12\linewidth}C{0.19\linewidth}C{0.19\linewidth}}
&\textit{``Vincent van Gogh''}&\textit{``Fauvism''}&&\textit{``Vincent van Gogh''}&\textit{``Fauvism''}
\end{tabular}
\includegraphics[width=1.0\linewidth]{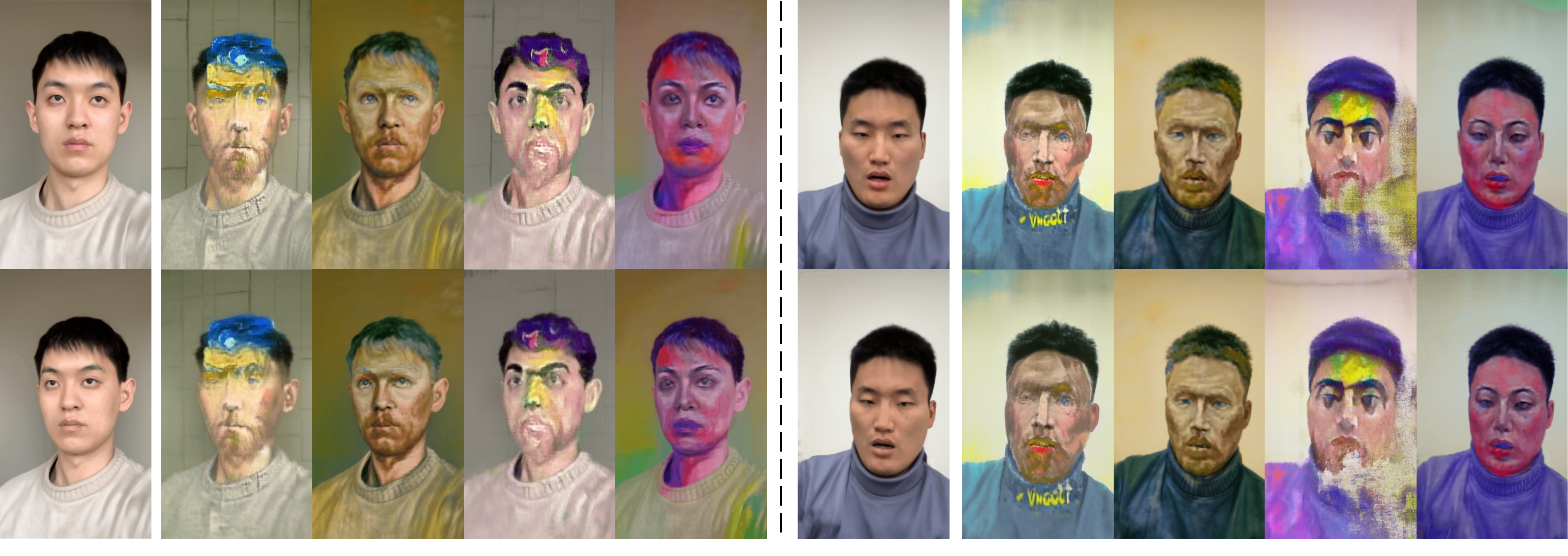}
\begin{tabular}{C{0.1\linewidth}C{0.1\linewidth}C{0.1\linewidth}C{0.1\linewidth}C{0.09\linewidth}C{0.01\linewidth}C{0.12\linewidth}C{0.09\linewidth}C{0.09\linewidth}C{0.11\linewidth}C{0.09\linewidth}}
Source&CLIP-NeRF&Ours&CLIP-NeRF&Ours&&Source&CLIP-NeRF&Ours&CLIP-NeRF&Ours\\&\\[-1.5ex]
\hline&\\[-1.5ex]
\end{tabular}
\begin{tabular}{C{0.49\linewidth}C{0.49\linewidth}}
\textit{``Checkerboard''}&\textit{``Mosaic Design''}
\end{tabular}
\includegraphics[width=1.0\linewidth]{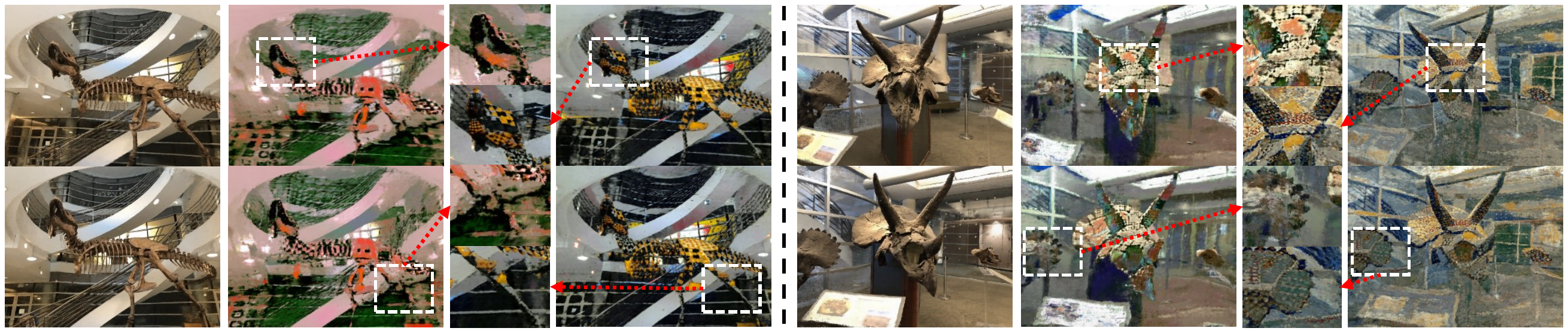}
\begin{tabular}{C{0.15\linewidth}C{0.14\linewidth}C{0.06\linewidth}C{0.15\linewidth}C{0.008\linewidth}C{0.15\linewidth}C{0.14\linewidth}C{0.06\linewidth}C{0.15\linewidth}}
Source&CLIP-NeRF&&Ours&&Source&CLIP-NeRF&&Ours\\&\\[-1.5ex]
\hline&\\[-1.5ex]
\end{tabular}
\begin{tabular}{C{0.12\linewidth}C{0.22\linewidth}C{0.2\linewidth}C{0.12\linewidth}C{0.18\linewidth}C{0.16\linewidth}}
&\textit{``Tolkien Elf''}&\textit{``Fauvism''}&&\textit{``Bat Man''}&\textit{``Hulk''}
\end{tabular}
\includegraphics[width=1.0\linewidth]{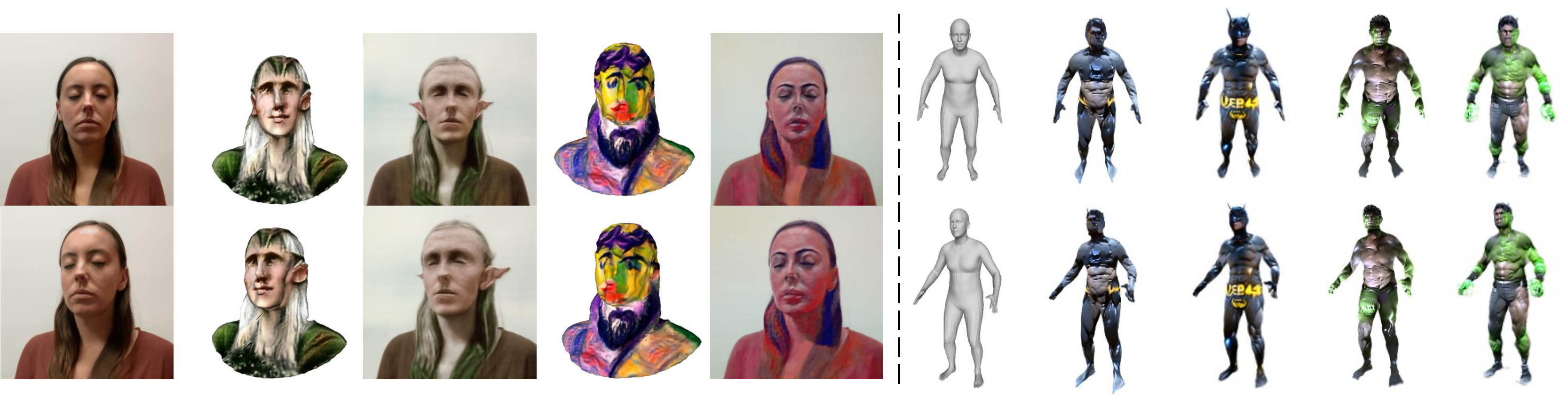}
\begin{tabular}{C{0.12\linewidth}C{0.11\linewidth}C{0.11\linewidth}C{0.11\linewidth}C{0.11\linewidth}C{0.11\linewidth}C{0.08\linewidth}C{0.08\linewidth}C{0.09\linewidth}C{0.08\linewidth}}
Source&DreamField&Ours&DreamField&Ours&Source&DreamField&Ours&DreamField&Ours\\&\\[-1.5ex]
\end{tabular}
\end{small}
\caption{\textbf{Comparisons.} Comparisons to text-guided NeRF stylization method \textit{CLIP-NeRF}~\cite{wang2021clip} and \textit{DreamField}~\cite{jain2022zero}. }
\label{fig:cmp2}
\end{figure*}

\begin{figure*}[t]
\centering
\setlength{\tabcolsep}{0\linewidth}
\begin{small}
\begin{tabular}{C{0.12\linewidth}C{0.22\linewidth}C{0.2\linewidth}C{0.12\linewidth}C{0.18\linewidth}C{0.16\linewidth}}
&\textit{``Vincent van Gogh''}&\textit{``Edvard Munch''}&&\textit{``Tolkien Elf''}&\textit{``Joker''}
\end{tabular}
\includegraphics[width=1.0\textwidth]{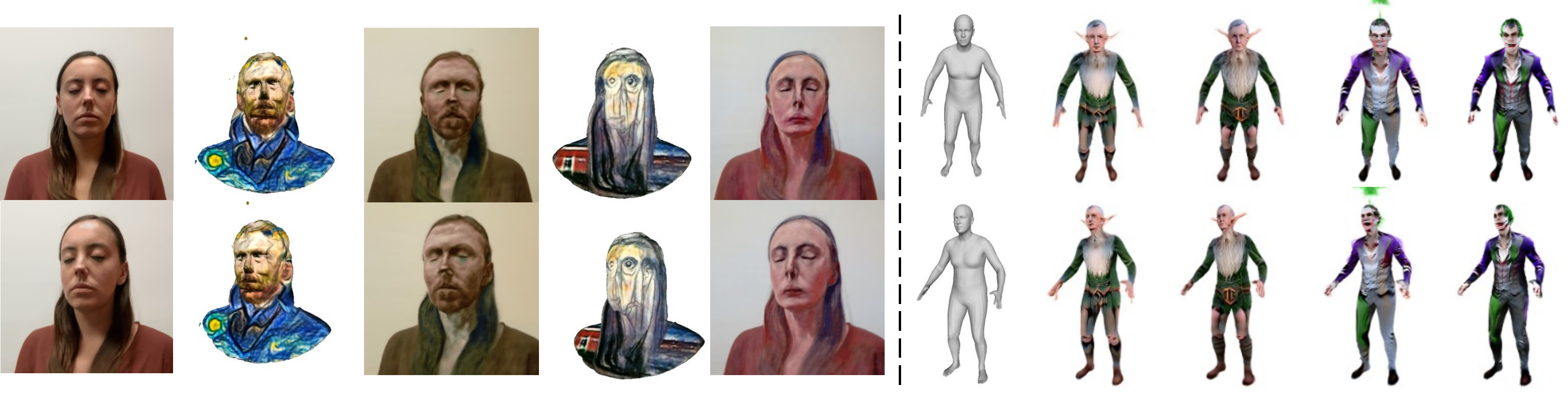}
\begin{tabular}{C{0.12\linewidth}C{0.11\linewidth}C{0.11\linewidth}C{0.11\linewidth}C{0.11\linewidth}C{0.11\linewidth}C{0.08\linewidth}C{0.08\linewidth}C{0.09\linewidth}C{0.08\linewidth}}
Source&AvatarCLIP&Ours&AvatarCLIP&Ours&Source&AvatarCLIP&Ours&AvatarCLIP&Ours\\&\\[-1.5ex]
\hline&\\[-1.5ex]
\end{tabular}
\begin{tabular}{C{0.12\linewidth}C{0.22\linewidth}C{0.2\linewidth}C{0.12\linewidth}C{0.18\linewidth}C{0.16\linewidth}}
&\textit{``Pixar''}&\textit{``Lord Voldemort''}&&\textit{``Iron Man''}&\textit{``Superman''}
\end{tabular}
\includegraphics[width=1.0\textwidth]{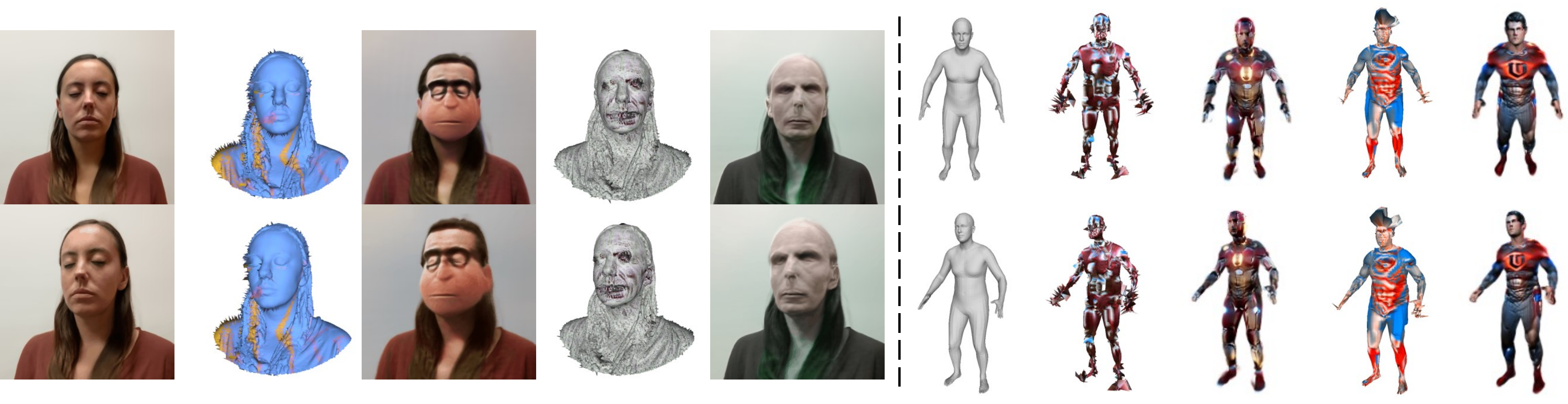}
\begin{tabular}{C{0.12\linewidth}C{0.11\linewidth}C{0.11\linewidth}C{0.11\linewidth}C{0.11\linewidth}C{0.11\linewidth}C{0.09\linewidth}C{0.07\linewidth}C{0.09\linewidth}C{0.08\linewidth}}
Source&Text2Mesh&Ours&Text2Mesh&Ours&Source&Text2Mesh&Ours&Text2Mesh&Ours\\[-1.5ex]
\end{tabular}
\end{small}
\caption{\textbf{Comparisons.} Comparisons to text-guided mesh-based stylization method \textit{Text2Mesh}~\cite{michel2021text2mesh} and \textit{AvatarCLIP}~\cite{Hong2022Avatar}.}
\label{fig:cmp3}
\end{figure*}

\begin{figure*}[th!]
\setlength{\tabcolsep}{0\linewidth}
\includegraphics[width=\textwidth]{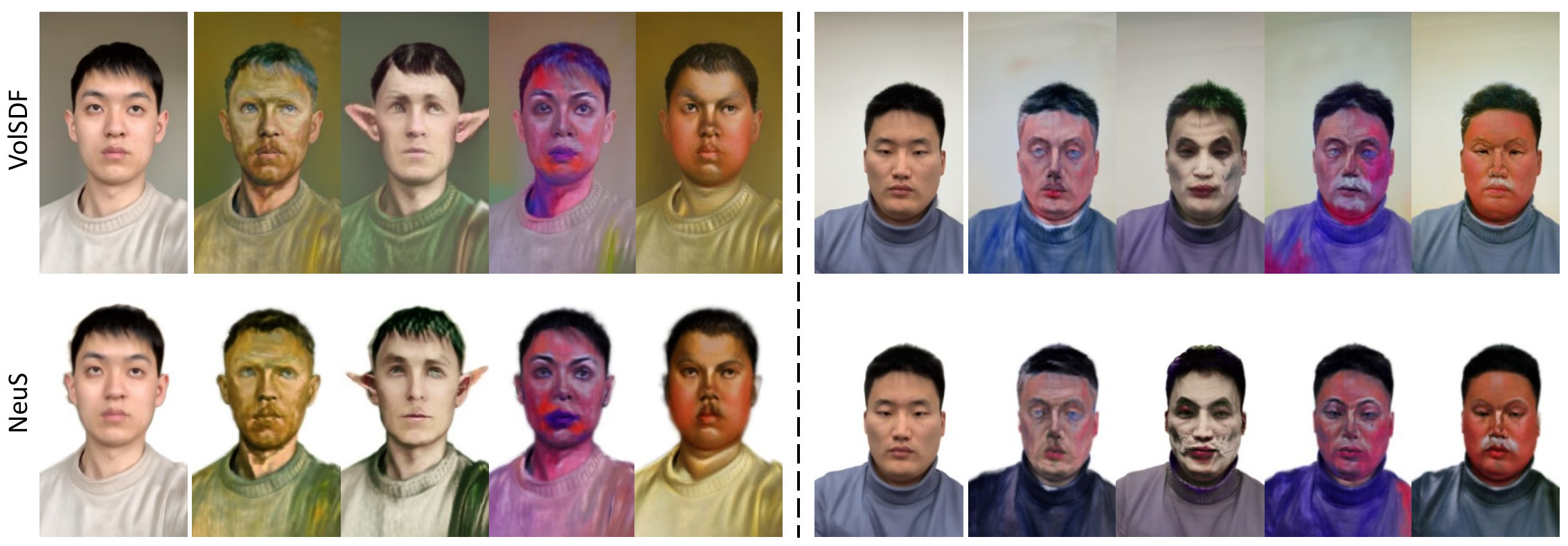}
\begin{small}
\begin{tabular}{C{0.02\linewidth}C{0.1\linewidth}C{0.1\linewidth}C{0.09\linewidth}C{0.09\linewidth}C{0.09\linewidth}C{0.03\linewidth}C{0.09\linewidth}C{0.1\linewidth}C{0.1\linewidth}C{0.09\linewidth}C{0.1\linewidth}}
&Source&\textit{``Vincent van Gogh''}&\textit{``Tolkien elf''}&\textit{``Fauvism''}&\textit{``Fernando Botero''}&&Source&\textit{``Edvard munch''}&\textit{``Joker''}&\textit{``Fauvism''}&\textit{``Fernando Botero''}
\end{tabular}
\end{small}
\caption{\textbf{Generalization Evaluation.} Generalization evaluation on VolSDF and NeuS.}
\label{fig:me}
\end{figure*}

\subsection{Text Evaluation}

As CLIP~\cite{park2020contrastive} is sensitive to text prompts, we conduct a text description evaluation in \Fref{fig:te}. 
When a text description refers to a style in general, not anyone in particular, the stylization can be insufficient. For example, \textit{``Fauvism''} only induces stylization around the mouth as it describes general meaning, like artists \textit{``Henri Matisse''} and \textit{``Kees van Dongen''} or \textit{``Brutalist painting''}. And the same observations when comparing \textit{``Chinese Painting''} and \textit{``Chinese Ink Painting''}.
In contrast, when a text refers to a specific object or style, the language ambiguity will disappear. For example, \textit{``Lord Voldemort''}, \textit{``Head of Lord Voldemort''}, and \textit{``Head of Lord Voldemort in fantasy style''} reveals similar stylization results. We also see the similar results concerning the Pixar style.
In the interests of brevity, we use \textit{``Fauvism''} to represent \textit{``painting, oil
on canvas, Fauvism style''} and \textit{``Vincent van Gogh''} to represent \textit{``painting, oil on canvas, Vincent van Gogh self-portrait style''} in other experiments. We also use the same prompt augmentation strategy for other painting styles, including \textit{``Edvard Munch''} and \textit{``Fernando Botero''}.

\begin{figure}[t]
\centering
\setlength{\tabcolsep}{0\linewidth}
\includegraphics[width=1.0\linewidth]{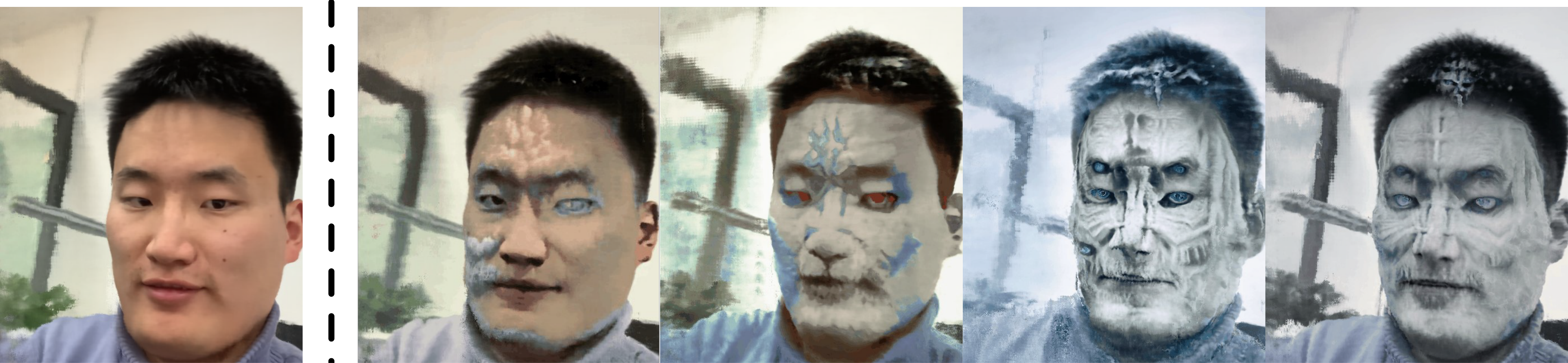}
\includegraphics[width=1.0\linewidth]{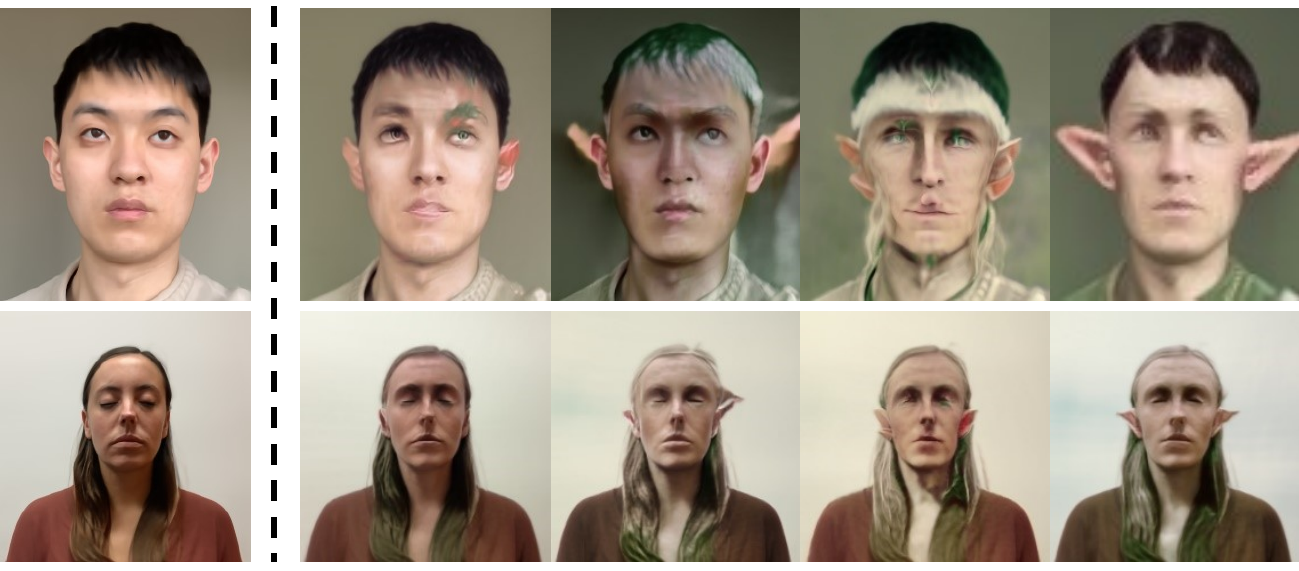}
\begin{small}
\begin{tabular}{C{0.18\linewidth}C{0.029\linewidth}C{0.2\linewidth}C{0.2\linewidth}C{0.2\linewidth}C{0.2\linewidth}}
Source&&w/o $\mathcal{L}_{con}^{g+l}$&w/o $\mathcal{L}_{con}^l$&w/o $\mathcal{L}_{con}^g$&Full
\end{tabular}
\end{small}
\caption{\textbf{Ablations on CLIP-Guided Losses.} Without our global-local contrastive losses, the results suffer from insufficient or non-uniform stylization. The target prompts are \textit{``White Walker''} and \textit{``Tolkien Elf''} respectively.}
\label{fig:ablation}
\end{figure}

\begin{figure*}[t]
\setlength{\tabcolsep}{0\linewidth}
\includegraphics[width=\textwidth]{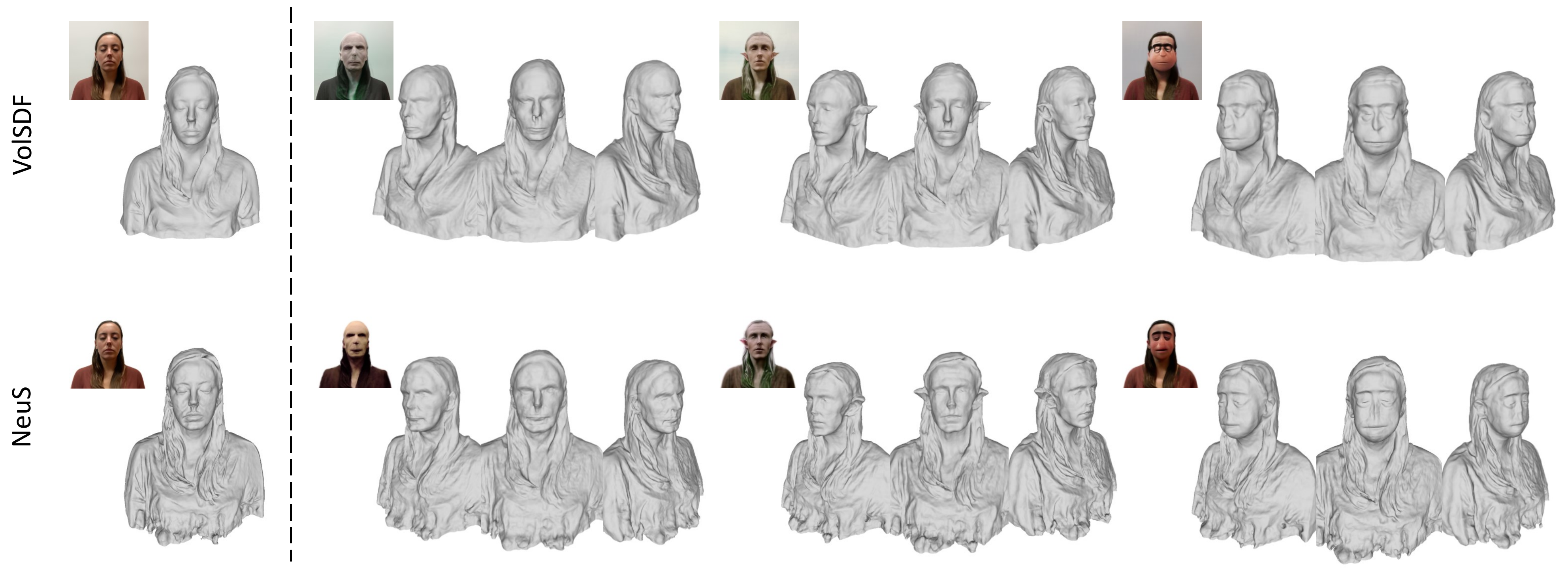}
\begin{small}
\begin{tabular}{C{0.188\linewidth}C{0.285\linewidth}C{0.280\linewidth}C{0.266\linewidth}}
Source&\textit{``Lord Voldemort''}&\textit{``Tolkien Elf''}&\textit{``Pixar''}
\end{tabular}
\end{small}
\caption{\textbf{Geometry Evaluation.} Our method modulates the geometry and color simultaneously of a pre-trained NeRF to match the desired style described by a text prompt.}
\label{fig:ge}
\end{figure*}

\begin{figure}[t]
\centering
\begin{small}
\begin{tabular}{C{\linewidth}}
\textit{``Disney 2D Superman''}
\end{tabular}
\end{small}
\setlength{\tabcolsep}{0\linewidth}
\includegraphics[width=1.0\linewidth]{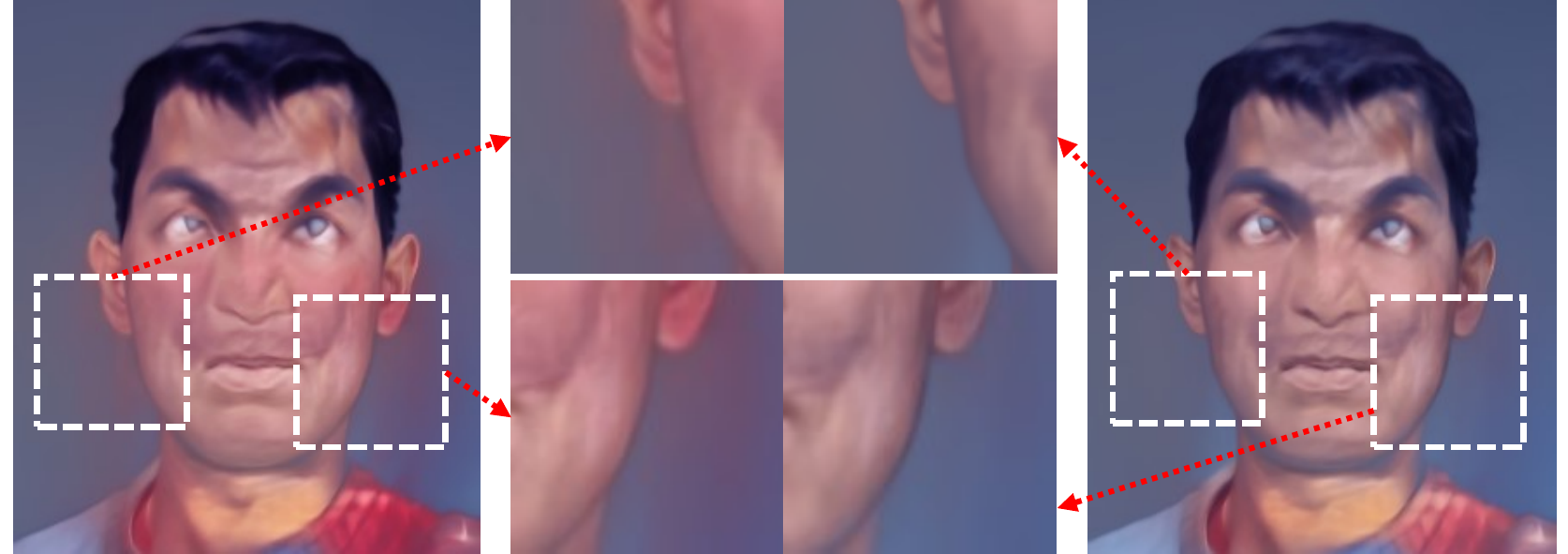}
\begin{small}
\begin{tabular}{C{\linewidth}}
\textit{``Groot, Guardians of the Galaxy''}
\end{tabular}
\end{small}
\includegraphics[width=0.98\linewidth]{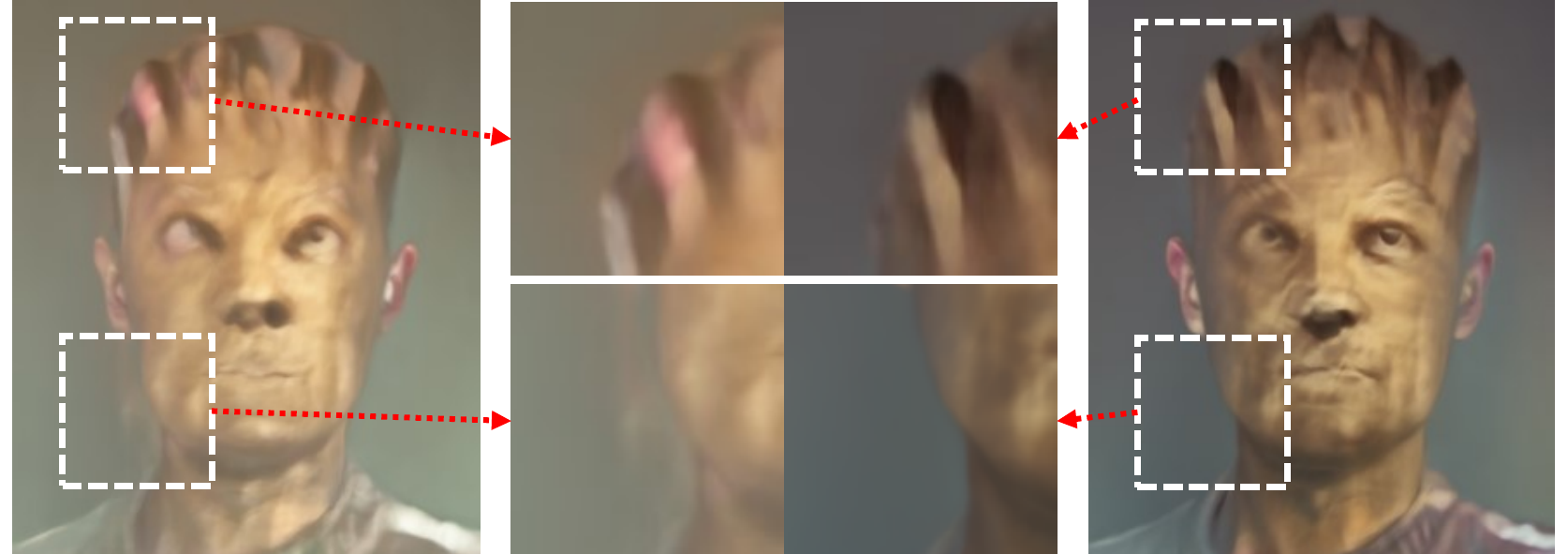}
\begin{small}
\begin{tabular}{C{0.33\linewidth}C{0.34\linewidth}C{0.33\linewidth}}
w/o $\mathcal{L}_{reg}$&&w/ $\mathcal{L}_{reg}$
\end{tabular}
\end{small}
\caption{\textbf{Ablations on Weight Regularization.} The cloudy artifacts near the corner or geometric noises are observed without the weight regularization loss.}
\label{fig:ablation_reg}
\end{figure}

\begin{figure}[t]
\centering
\includegraphics[width=0.98\linewidth]{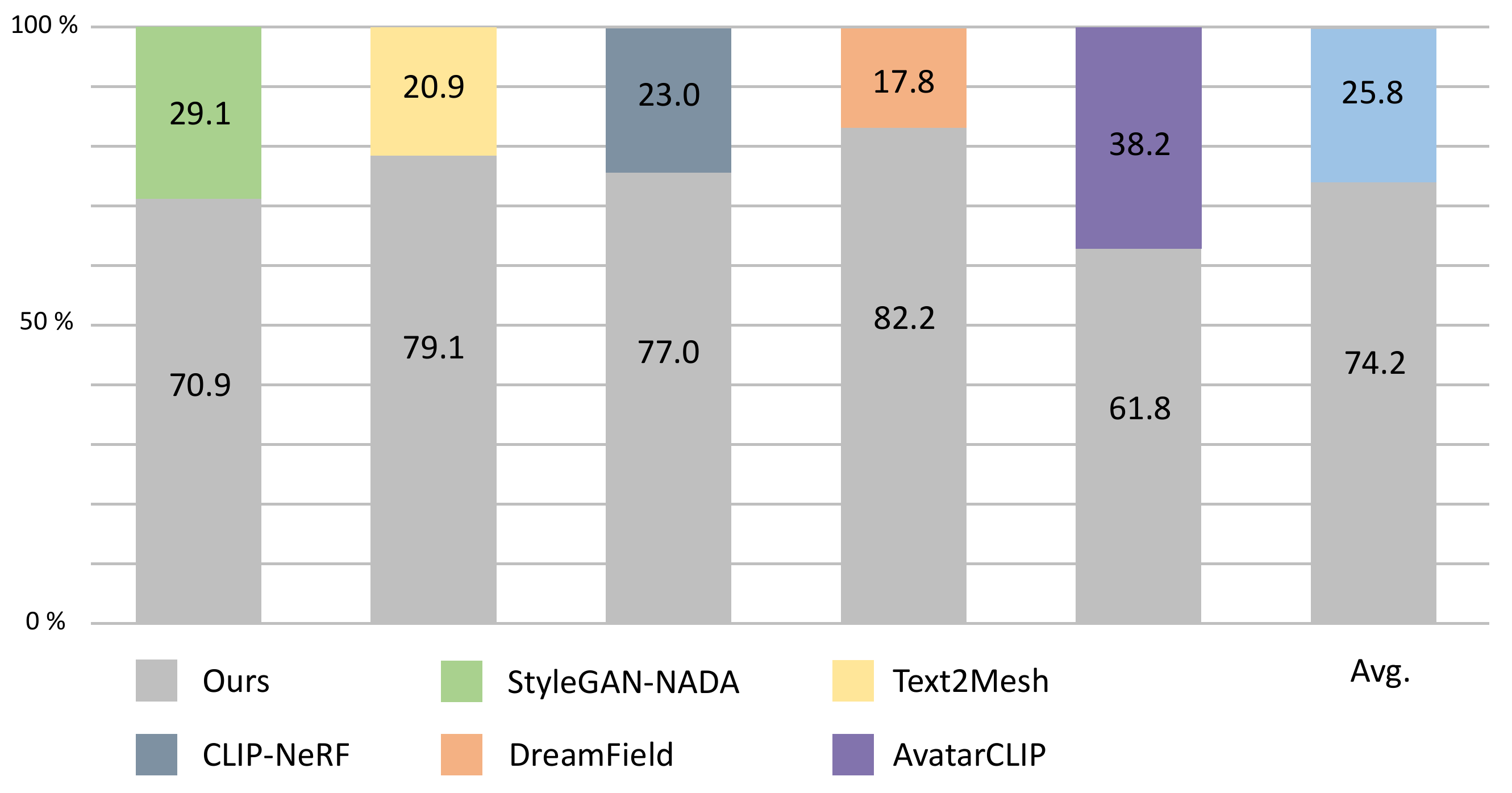}
\caption{\textbf{User Study.} Our method consistently outperforms state-of-the-art text-guided stylization methods on user preference rates~(\%).}
\label{fig:user_study}
\end{figure}

\subsection{Comparisons}

We compare with most related works following three categories: 1) Text-driven image stylization: \textit{StyleGAN-NADA}~\cite{gal2021stylegan}; 2) Text-driven mesh-based stylization: \textit{Text2Mesh}~\cite{michel2021text2mesh} and \textit{AvatarCLIP}~\cite{Hong2022Avatar}; and 3) Text-driven NeRF stylization: \textit{CLIP-NeRF}~\cite{wang2021clip} and \textit{DreamField}~\cite{jain2022zero}.
To make fair comparisons with these methods, we adopt author-released codes and accommodate the input to each method as required. For \textit{StyleGAN-NADA}, we follow its steps to first conduct a face alignment under the setting of \textit{FFHQ}~\cite{karras2019style} and then invert these faces using \textit{e4e}~\cite{tov2021designing} into latent codes, before inputting them to \textit{StyleGAN-NADA}. We have also tried \textit{pSp}~\cite{richardson2021encoding} to invert latent codes but finally adopt \textit{e4e} to obtain better stylization results. Per the authors' advice, we trained 600 iterations and sampled faces present visual-pleasing stylized results. We place final stylized faces back on the input images by inversing the face alignment process.
As for \textit{Text2Mesh}, the input mesh of one example (\textit{`Lady'}) is provided by the \textit{H3DS}~\cite{ramon2021h3d}, while the input mesh of another example (\textit{`Human'}) is fetched from \textit{AvatarCLIP}. Both meshes are normalized into -1 to 1, before inputting them to \textit{Text2Mesh}. We follow the training setting of \textit{Text2Mesh} in stylizing the person object to stylize \textit{`Lady'} and  \textit{`Human'}.
We compare to \textit{DreamField} and \textit{AvatarCLIP} following the shape sculpting and texture generation process of \textit{AvatarCLIP}. Similar to \textit{AvatarCLIP}, we also adopt prompt augmentations when stylizing the \textit{`Human'}. For example, we use text prompts including \textit{``Tolkien Elf''}, \textit{``the back of Tolkien Elf''}, and \textit{``the face of Tolkien Elf''} for the detailed refinement. 

The visual comparisons are demonstrated in \Fref{fig:cmp1}, \Fref{fig:cmp2}, and \Fref{fig:cmp3}. For video results, please see the supplementary material.

\noindent\textbf{\emph{Comparisons to text-driven image stylization.}} Compared to \textit{StyleGAN-NADA}, our method can better ensure the desired style strength in all examples by introducing global-local contrastive learning. \textit{StyleGAN-NADA} achieves visual-pleasing results on sampled faces but reflects a degradation for in-the-wild faces partly due to the latent code inversion. 
Moreover, as a 3D stylization method, ours can preserve view consistencies in the stylized results. In contrast, \textit{StyleGAN-NADA} stylizes each view independently, thus introducing inconsistent shapes or textures to different views. This may lead to flickering artifacts when applied to video applications. Moreover, \textit{StyleGAN-NADA} is less friendly to real faces as the input image has to be inverted back to the StyleGAN latent space before stylization, which will inevitably lead to some detail loss and identity change. Unlike it, \textit{NeRF-Art} is not constrained by any latent space of pre-trained networks and does not need the inversion step. 

\noindent\textbf{\emph{Comparisons to text-driven NeRF stylization.}}
Compared with \textit{CLIP-NeRF}, our advantages are two-fold. First, \textit{CLIP-NeRF} stylizes NeRF using the absolute directional loss, which does not put enough stylizations. Moreover, it suffers from uneven stylizations. For example, we only see enough stylizations on the nose and hair for style \textit{``Fauvism''}, but the man's cheek has not been fully stylized. In contrast,
we design a global-local contrastive learning strategy to ensure the desired style strength. Second, as no weight regularization is used in \textit{CLIP-NeRF}, its results may appear as severe geometry noises. In contrast, our weight regularization suppresses geometric noises by encouraging a more concentrated density distribution.
\textit{DreamField} also adopts the absolute directional loss to stylize NeRF, which cannot guarantee sufficient and uniform stylization. 
\textit{DreamField} adopts a random background augmentation to CLIP's attention on the foreground, which requires view-consistent masks, while ours does not. Moreover, our method consistently outperforms \textit{DreamField} in detailed cloth wrinkles, facial attributes, and fine-grained geometry deformations, like muscle shapes and antennas.
In summary, our \textit{NeRF-Art} outperforms these methods by proposing a contrastive learning technique to achieve sufficient and uniform stylization and 
designing a weight regularization to remove cloudy artifacts and geometry noises.

\noindent\textbf{\emph{Comparisons to text-driven mesh-based stylization.}}
\textit{Text2Mesh} also supports geometry deformation and texture stylization of a 3D model like ours. However, it assumes there exists a synergy between the input 3D geometry and the target prompt and is more likely to fail when stylizing a 3D mesh towards a less related prompt, such as \textit{``Pixar''} for the \textit{`Lady'} model in \Fref{fig:cmp2}. With carefully-designed loss constraints, ours is more robust to different prompts, either related to the 3D scenes or not. 
Moreover, limited by the expressivity of the mesh representation, \textit{Text2Mesh} fails most runs and presents unstable stylization results, resulting in irregular deformations and indentations on the edge or surface. Authors of \textit{AvatarCLIP} also report similar results when comparing to \textit{Text2Mesh}. 
Similar to \textit{DreamField}, \textit{AvatarCLIP} adopts a random background augmentation to lead CLIP to focus on the foreground and prevents floating artifact generations. Nevertheless, this process requires view-consistent masks while ours does not. Moreover, \textit{AvatarCLIP} adds an additional color network to constrain the general shape of the avatar as well as introducing random shading and lighting augmentations on the textured renderings to strengthen the stylization. Even with these augmentations, \textit{AvatarCLIP} still fails to produce satisfying texture and geometry details. In contrast, ours reveals a fine-grained beard, detailed wrinkles of garments, and clearer face attributes. Noteworthy, our \textit{NeRF-Art} supports stylizing in-the-wild faces, while \textit{AvatarCLIP} requires a 3D mesh as input to conduct these augmentations. 
Finally, \textit{AvatarCLIP} can still generate random bumps in the background and make the extracted surface noisy. This is because \textit{AvatarCLIP} sampled a sparse rays~($112\times112$) to construct a coarse renderings for CLIP constraints, due to OOM problem. We found worse results with more noise when reducing sampled ray numbers. In contrast, our method supports training stylization on all rays by imposing a memory-saving technique. In conclusion, \textit{NeRF-Art} achieves better stylization using the proposed contrastive learning strategies without any mesh guidance. 

\subsection{User Study}
To evaluate stylization quality from human perception, we conducted a user study. For each compared category, we used two subjects. For each subject, we selected $5$ prompts from our text descriptions dataset and finally obtained $10$ test cases for each category and $50$ in total. For every test case, we showed one sample of input frames, the textual prompt, and the results of different methods in two views and random order. The participants were given unlimited time to select the best stylization results by jointly considering three aspects: preservation of the content, faithfulness to the style, and view consistency. We finally collected $23$ questionnaires completed by $10$ male and $13$ Lady participants. Statistics of the user study are shown in \Fref{fig:user_study}. Our method outperforms \textit{StyleGAN-NADA}, \textit{CLIP-NeRF}, \textit{Text2Mesh}, \textit{DreamField}, and \textit{AvatarCLIP} by achieving much higher user preference rates.

\subsection{Ablation Study}
\label{subsec:Ablation}

\noindent\textbf{\emph{Why global-local contrastive learning?}}
A straightforward way to stylize NeRFs is to apply the directional CLIP loss proposed by StyleGAN-NADA~\cite{gal2021stylegan} to the rendered views. Unfortunately, the directional CLIP loss can enforce the right stylization trajectory but struggles to reach a sufficient magnitude, as shown in the 2nd column of \Fref{fig:ablation}. This is because the loss only measures the directional similarity between the normalized embedded vectors but ignores their actual distances. In contrast, our global contrastive loss (3rd column of \Fref{fig:ablation}) can ensure the proper stylization magnitude by pushing it as close as possible to the target. However, the global contrastive loss still cannot guarantee a sufficient and uniform stylization of the whole scene. 
The stylization shows excess on certain parts and insufficiency on others, e.g., insufficient stylized faces and excessively stylized eyes in the \textit{``Tolkien Elf''} example in the 3rd column of \Fref{fig:ablation}. This may attribute to the fact that CLIP focuses more attention on regions with distinguishable features than on other regions.
Our local contrastive loss helps achieve more balanced stylized results by stylizing every local region of the scene (4th column of \Fref{fig:ablation}). However, this local contrastive loss without global information may produce excessive facial attributes, e.g., generating more eyes in the \textit{``White Walker''} example and two left ears in the \textit{``Tolkien Elf''} example. This attributes to insufficient semantics involved in a local patch. This problem can be avoided by adding the global contrastive loss at the same time.

By combining both global and local contrastive loss with the directional CLIP, our method successfully achieves uniform stylization with both correct stylization direction and sufficient magnitude (5th column of \Fref{fig:ablation}).

\noindent\textbf{\emph{Why weight regularization?}}
Altering the geometry of NeRF may potentially cause cloudy artifacts. 
In \Fref{fig:ablation_reg}, we demonstrate that the weight regularization loss can suppress cloudy artifacts and geometric noises by encouraging a more concentrated density distribution for stylization.

\subsection{Generalization Evaluation}

We conduct a generalization evaluation on VolSDF and NeuS in \Fref{fig:me} to evaluate \textit{NeRF-Art}'s ability in adapting to different NeRF-like models. As NeuS reconstructs a coarse result on our in-the-wild data without a mask may due to inaccurate camera estimations, we conduct a segmentation using RVM~\cite{lin2022robust} for better reconstruction and dilate the mask using OpenCV with $3\times3$ kernel and two iterations to allow geometry variations. In \Fref{fig:me}, our method presents similar stylization results on VolSDF and NeuS, which demonstrates that our \textit{NeRF-Art} has the ability to adapt to different NeRF-like models.

\subsection{Geometry Evaluation}

To evaluate whether the geometry will be correctly modulated in the stylization process, we show the geometry evaluation results in \Fref{fig:ge}. We extract meshes using Marching Cubes~\cite{lorensen1987marching} before and after the stylization for comparison and report results on two widely-used NeRF-like models VolSDF~\cite{yariv2021volume} and NeuS~\cite{wang2021neus}.
We clearly see geometry changes by comparison with the source mesh. For example, \textit{``Lord Voldemort''}  flattens the girl's nose, \textit{``Tolkien Elf''} sharpens the girl's ears, and \textit{``Pixar''} rounds the jaw. Moreover, we find the same observations on both VolSDF and NeuS. In summary, we conclude that our method can correctly modulate the geometry of NeRF to match the desired style. 

\begin{figure}[t]
\centering
\setlength{\tabcolsep}{0\linewidth}
\includegraphics[width=0.92\linewidth]{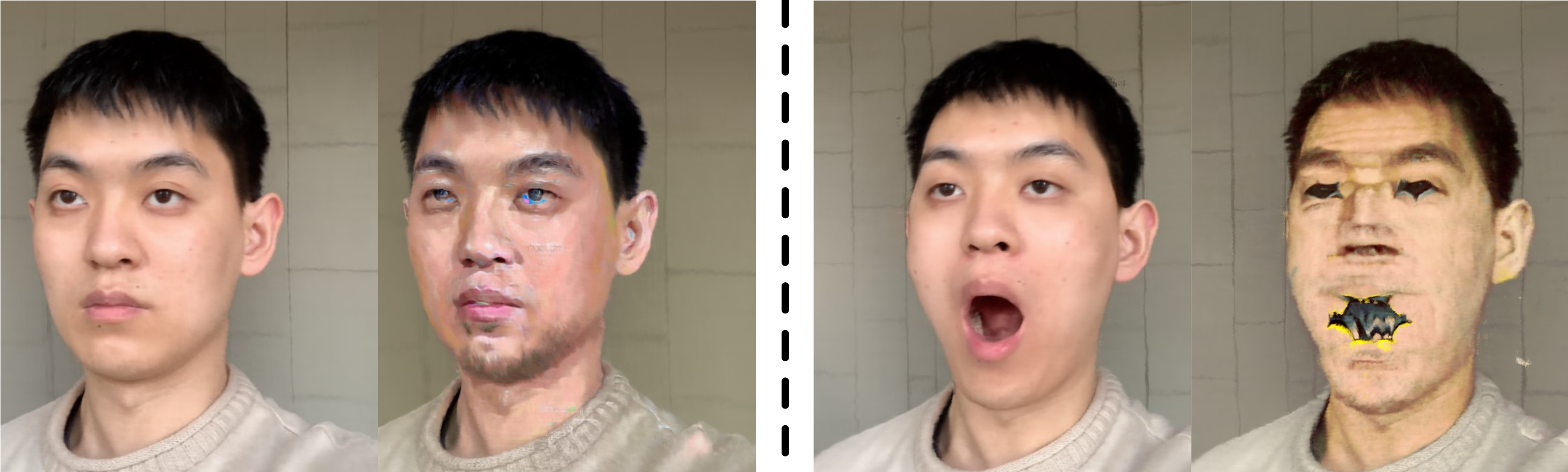}
\begin{small}
\begin{tabular}{C{0.2\linewidth}C{0.24\linewidth}C{0.04\linewidth}C{0.22\linewidth}C{0.22\linewidth}}
Source&\textit{``Digital painting''}&&Source&\textit{``Mouth Batman''}
\end{tabular}
\end{small}
\caption{\textbf{Limitations.} Linguistic ambiguity (left) or semantically meaningless words (right) may lead to unexpected results.}
\label{fig:limit}
\end{figure}

\section{Conclusion}

In this paper, we present \textit{NeRF-Art}, the text-guided NeRF stylization approach based on CLIP. Unlike existing approaches that require the mesh guidance in the stylization process or traps in insufficient geometry deformations and texture details in stylization,
ours modulate its geometry and appearance simultaneously to match the desired style and show visual-pleasing results of geometry deformations and texture details with only a text guidance.
To achieve it, we introduce a carefully-designed combination of directional constraint to control the style trajectory and novel global-local contrastive loss to enforce the proper style strength. Moreover, we propose a weight regularization strategy to alleviate the cloudy artifacts and geometry noises in deforming the geometry. Extensive experiments on real faces and general scenes show that our method is effective and robust in both stylization quality and view consistency.

\noindent{\textbf{Limitations.}} 
Despite the success in most cases, our method still has some limitations. First, some text prompts are linguistically ambiguous, like \textit{``Digital painting''}, which describes a wide range of styles, including oil paintings, pencil sketches, 3D rendering images, cartoon drawings, etc. This ambiguity might confuse the CLIP and make the final result unexpected, as shown in \Fref{fig:limit}. Semantically meaningless words cause another kind of unexpected result. For example, if we combine the words \textit{``Mouth''} and \textit{``Batman''} as a prompt, the result unexpectedly puts a bat shape on the mouth, which may not be what the user desires. These are interesting problems worth exploring in the future.

% Bibliography
\bibliographystyle{ACM-Reference-Format}
\bibliography{ref}


\begin{thebibliography}{94}

%%% the Web address.

\ifx \showCODEN    \undefined \def \showCODEN     #1{\unskip}     \fi
\ifx \showDOI      \undefined \def \showDOI       #1{#1}\fi
\ifx \showISBNx    \undefined \def \showISBNx     #1{\unskip}     \fi
\ifx \showISBNxiii \undefined \def \showISBNxiii  #1{\unskip}     \fi
\ifx \showISSN     \undefined \def \showISSN      #1{\unskip}     \fi
\ifx \showLCCN     \undefined \def \showLCCN      #1{\unskip}     \fi
\ifx \shownote     \undefined \def \shownote      #1{#1}          \fi
\ifx \showarticletitle \undefined \def \showarticletitle #1{#1}   \fi
\ifx \showURL      \undefined \def \showURL       {\relax}        \fi
% invisible to TeX
\providecommand\bibfield[2]{#2}
\providecommand\bibinfo[2]{#2}
\providecommand\natexlab[1]{#1}
\providecommand\showeprint[2][]{arXiv:#2}

\bibitem[Arandjelovi{\'c} and Zisserman(2021)]%
        {arandjelovic2021nerf}
\bibfield{author}{\bibinfo{person}{Relja Arandjelovi{\'c}} {and}
  \bibinfo{person}{Andrew Zisserman}.} \bibinfo{year}{2021}\natexlab{}.
\newblock \showarticletitle{Nerf in detail: Learning to sample for view
  synthesis}.
\newblock \bibinfo{journal}{\emph{arXiv preprint arXiv:2106.05264}}
  (\bibinfo{year}{2021}).
\newblock


\bibitem[Barron et~al\mbox{.}(2021)]%
        {barron2021mip}
\bibfield{author}{\bibinfo{person}{Jonathan~T Barron}, \bibinfo{person}{Ben
  Mildenhall}, \bibinfo{person}{Matthew Tancik}, \bibinfo{person}{Peter
  Hedman}, \bibinfo{person}{Ricardo Martin-Brualla}, {and}
  \bibinfo{person}{Pratul~P Srinivasan}.} \bibinfo{year}{2021}\natexlab{}.
\newblock \showarticletitle{Mip-nerf: A multiscale representation for
  anti-aliasing neural radiance fields}. In
  \bibinfo{booktitle}{\emph{Proceedings of the IEEE/CVF International
  Conference on Computer Vision}}. \bibinfo{pages}{5855--5864}.
\newblock


\bibitem[Barron et~al\mbox{.}(2022)]%
        {barron2022mipnerf360}
\bibfield{author}{\bibinfo{person}{Jonathan~T. Barron}, \bibinfo{person}{Ben
  Mildenhall}, \bibinfo{person}{Dor Verbin}, \bibinfo{person}{Pratul~P.
  Srinivasan}, {and} \bibinfo{person}{Peter Hedman}.}
  \bibinfo{year}{2022}\natexlab{}.
\newblock \showarticletitle{Mip-NeRF 360: Unbounded Anti-Aliased Neural
  Radiance Fields}.
\newblock \bibinfo{journal}{\emph{CVPR}} (\bibinfo{year}{2022}).
\newblock


\bibitem[Cao et~al\mbox{.}(2020)]%
        {cao2020psnet}
\bibfield{author}{\bibinfo{person}{Xu Cao}, \bibinfo{person}{Weimin Wang},
  \bibinfo{person}{Katashi Nagao}, {and} \bibinfo{person}{Ryosuke Nakamura}.}
  \bibinfo{year}{2020}\natexlab{}.
\newblock \showarticletitle{Psnet: A style transfer network for point cloud
  stylization on geometry and color}. In \bibinfo{booktitle}{\emph{Proceedings
  of the IEEE/CVF Winter Conference on Applications of Computer Vision}}.
  \bibinfo{pages}{3337--3345}.
\newblock


\bibitem[Chen et~al\mbox{.}(2017a)]%
        {chen2017coherent}
\bibfield{author}{\bibinfo{person}{Dongdong Chen}, \bibinfo{person}{Jing Liao},
  \bibinfo{person}{Lu Yuan}, \bibinfo{person}{Nenghai Yu}, {and}
  \bibinfo{person}{Gang Hua}.} \bibinfo{year}{2017}\natexlab{a}.
\newblock \showarticletitle{Coherent online video style transfer}. In
  \bibinfo{booktitle}{\emph{Proceedings of the IEEE International Conference on
  Computer Vision}}. \bibinfo{pages}{1105--1114}.
\newblock


\bibitem[Chen et~al\mbox{.}(2017b)]%
        {chen2017stylebank}
\bibfield{author}{\bibinfo{person}{Dongdong Chen}, \bibinfo{person}{Lu Yuan},
  \bibinfo{person}{Jing Liao}, \bibinfo{person}{Nenghai Yu}, {and}
  \bibinfo{person}{Gang Hua}.} \bibinfo{year}{2017}\natexlab{b}.
\newblock \showarticletitle{Stylebank: An explicit representation for neural
  image style transfer}. In \bibinfo{booktitle}{\emph{Proceedings of the IEEE
  Conference on Computer Vision and Pattern Recognition}}.
  \bibinfo{pages}{1897--1906}.
\newblock


\bibitem[Chen et~al\mbox{.}(2018)]%
        {chen2018stereoscopic}
\bibfield{author}{\bibinfo{person}{Dongdong Chen}, \bibinfo{person}{Lu Yuan},
  \bibinfo{person}{Jing Liao}, \bibinfo{person}{Nenghai Yu}, {and}
  \bibinfo{person}{Gang Hua}.} \bibinfo{year}{2018}\natexlab{}.
\newblock \showarticletitle{Stereoscopic neural style transfer}. In
  \bibinfo{booktitle}{\emph{Proceedings of the IEEE Conference on Computer
  Vision and Pattern Recognition}}. \bibinfo{pages}{6654--6663}.
\newblock


\bibitem[Chen et~al\mbox{.}(2020)]%
        {chen2020optical}
\bibfield{author}{\bibinfo{person}{Xinghao Chen}, \bibinfo{person}{Yiman
  Zhang}, \bibinfo{person}{Yunhe Wang}, \bibinfo{person}{Han Shu},
  \bibinfo{person}{Chunjing Xu}, {and} \bibinfo{person}{Chang Xu}.}
  \bibinfo{year}{2020}\natexlab{}.
\newblock \showarticletitle{Optical flow distillation: Towards efficient and
  stable video style transfer}. In \bibinfo{booktitle}{\emph{European
  Conference on Computer Vision}}. Springer, \bibinfo{pages}{614--630}.
\newblock


\bibitem[Chiang et~al\mbox{.}(2022)]%
        {chiang2022stylizing}
\bibfield{author}{\bibinfo{person}{Pei-Ze Chiang}, \bibinfo{person}{Meng-Shiun
  Tsai}, \bibinfo{person}{Hung-Yu Tseng}, \bibinfo{person}{Wei-Sheng Lai},
  {and} \bibinfo{person}{Wei-Chen Chiu}.} \bibinfo{year}{2022}\natexlab{}.
\newblock \showarticletitle{Stylizing 3D Scene via Implicit Representation and
  HyperNetwork}. In \bibinfo{booktitle}{\emph{Proceedings of the IEEE/CVF
  Winter Conference on Applications of Computer Vision}}.
  \bibinfo{pages}{1475--1484}.
\newblock


\bibitem[Deng et~al\mbox{.}(2021)]%
        {deng2021depth}
\bibfield{author}{\bibinfo{person}{Kangle Deng}, \bibinfo{person}{Andrew Liu},
  \bibinfo{person}{Jun-Yan Zhu}, {and} \bibinfo{person}{Deva Ramanan}.}
  \bibinfo{year}{2021}\natexlab{}.
\newblock \showarticletitle{Depth-supervised nerf: Fewer views and faster
  training for free}.
\newblock \bibinfo{journal}{\emph{arXiv preprint arXiv:2107.02791}}
  (\bibinfo{year}{2021}).
\newblock


\bibitem[Deng et~al\mbox{.}(2022)]%
        {deng2022fov}
\bibfield{author}{\bibinfo{person}{Nianchen Deng}, \bibinfo{person}{Zhenyi He},
  \bibinfo{person}{Jiannan Ye}, \bibinfo{person}{Budmonde Duinkharjav},
  \bibinfo{person}{Praneeth Chakravarthula}, \bibinfo{person}{Xubo Yang}, {and}
  \bibinfo{person}{Qi Sun}.} \bibinfo{year}{2022}\natexlab{}.
\newblock \showarticletitle{FoV-NeRF: Foveated Neural Radiance Fields for
  Virtual Reality}.
\newblock \bibinfo{journal}{\emph{IEEE Transactions on Visualization and
  Computer Graphics}} \bibinfo{volume}{28}, \bibinfo{number}{11}
  (\bibinfo{year}{2022}), \bibinfo{pages}{3854--3864}.
\newblock


\bibitem[Fan et~al\mbox{.}(2022)]%
        {fan2022unified}
\bibfield{author}{\bibinfo{person}{Zhiwen Fan}, \bibinfo{person}{Yifan Jiang},
  \bibinfo{person}{Peihao Wang}, \bibinfo{person}{Xinyu Gong},
  \bibinfo{person}{Dejia Xu}, {and} \bibinfo{person}{Zhangyang Wang}.}
  \bibinfo{year}{2022}\natexlab{}.
\newblock \showarticletitle{Unified Implicit Neural Stylization}.
\newblock \bibinfo{journal}{\emph{arXiv preprint arXiv:2204.01943}}
  (\bibinfo{year}{2022}).
\newblock


\bibitem[Gal et~al\mbox{.}(2021)]%
        {gal2021stylegan}
\bibfield{author}{\bibinfo{person}{Rinon Gal}, \bibinfo{person}{Or Patashnik},
  \bibinfo{person}{Haggai Maron}, \bibinfo{person}{Gal Chechik}, {and}
  \bibinfo{person}{Daniel Cohen-Or}.} \bibinfo{year}{2021}\natexlab{}.
\newblock \showarticletitle{Stylegan-nada: Clip-guided domain adaptation of
  image generators}.
\newblock \bibinfo{journal}{\emph{arXiv preprint arXiv:2108.00946}}
  (\bibinfo{year}{2021}).
\newblock


\bibitem[Gao et~al\mbox{.}(2021)]%
        {gao2021dynamic}
\bibfield{author}{\bibinfo{person}{Chen Gao}, \bibinfo{person}{Ayush Saraf},
  \bibinfo{person}{Johannes Kopf}, {and} \bibinfo{person}{Jia-Bin Huang}.}
  \bibinfo{year}{2021}\natexlab{}.
\newblock \showarticletitle{Dynamic view synthesis from dynamic monocular
  video}. In \bibinfo{booktitle}{\emph{Proceedings of the IEEE/CVF
  International Conference on Computer Vision}}. \bibinfo{pages}{5712--5721}.
\newblock


\bibitem[Garbin et~al\mbox{.}(2021)]%
        {garbin2021fastnerf}
\bibfield{author}{\bibinfo{person}{Stephan~J Garbin}, \bibinfo{person}{Marek
  Kowalski}, \bibinfo{person}{Matthew Johnson}, \bibinfo{person}{Jamie
  Shotton}, {and} \bibinfo{person}{Julien Valentin}.}
  \bibinfo{year}{2021}\natexlab{}.
\newblock \showarticletitle{Fastnerf: High-fidelity neural rendering at
  200fps}. In \bibinfo{booktitle}{\emph{Proceedings of the IEEE/CVF
  International Conference on Computer Vision}}. \bibinfo{pages}{14346--14355}.
\newblock


\bibitem[Gatys et~al\mbox{.}(2016)]%
        {gatys2016image}
\bibfield{author}{\bibinfo{person}{Leon~A Gatys}, \bibinfo{person}{Alexander~S
  Ecker}, {and} \bibinfo{person}{Matthias Bethge}.}
  \bibinfo{year}{2016}\natexlab{}.
\newblock \showarticletitle{Image style transfer using convolutional neural
  networks}. In \bibinfo{booktitle}{\emph{Proceedings of the IEEE conference on
  computer vision and pattern recognition}}. \bibinfo{pages}{2414--2423}.
\newblock


\bibitem[Guo et~al\mbox{.}(2021)]%
        {guo2021volumetric}
\bibfield{author}{\bibinfo{person}{Jie Guo}, \bibinfo{person}{Mengtian Li},
  \bibinfo{person}{Zijing Zong}, \bibinfo{person}{Yuntao Liu},
  \bibinfo{person}{Jingwu He}, \bibinfo{person}{Yanwen Guo}, {and}
  \bibinfo{person}{Ling-Qi Yan}.} \bibinfo{year}{2021}\natexlab{}.
\newblock \showarticletitle{Volumetric appearance stylization with stylizing
  kernel prediction network}.
\newblock \bibinfo{journal}{\emph{ACM Trans Graph}}  \bibinfo{volume}{40}
  (\bibinfo{year}{2021}), \bibinfo{pages}{1--15}.
\newblock


\bibitem[Han et~al\mbox{.}(2021a)]%
        {han2021exemplarbased}
\bibfield{author}{\bibinfo{person}{Fangzhou Han}, \bibinfo{person}{Shuquan Ye},
  \bibinfo{person}{Mingming He}, \bibinfo{person}{Menglei Chai}, {and}
  \bibinfo{person}{Jing Liao}.} \bibinfo{year}{2021}\natexlab{a}.
\newblock \showarticletitle{Exemplar-Based 3D Portrait Stylization}.
\newblock \bibinfo{journal}{\emph{IEEE Transactions on Visualization and
  Computer Graphics}} (\bibinfo{year}{2021}).
\newblock
\urldef\tempurl%
\url{https://doi.org/10.1109/TVCG.2021.3114308}
\showDOI{\tempurl}


\bibitem[Han et~al\mbox{.}(2021b)]%
        {han2021exemplar}
\bibfield{author}{\bibinfo{person}{Fangzhou Han}, \bibinfo{person}{Shuquan Ye},
  \bibinfo{person}{Mingming He}, \bibinfo{person}{Menglei Chai}, {and}
  \bibinfo{person}{Jing Liao}.} \bibinfo{year}{2021}\natexlab{b}.
\newblock \showarticletitle{Exemplar-based 3d portrait stylization}.
\newblock \bibinfo{journal}{\emph{IEEE Transactions on Visualization and
  Computer Graphics}} (\bibinfo{year}{2021}).
\newblock


\bibitem[Hertzmann(1998)]%
        {hertzmann1998painterly}
\bibfield{author}{\bibinfo{person}{Aaron Hertzmann}.}
  \bibinfo{year}{1998}\natexlab{}.
\newblock \showarticletitle{Painterly rendering with curved brush strokes of
  multiple sizes}. In \bibinfo{booktitle}{\emph{Proceedings of the 25th annual
  conference on Computer graphics and interactive techniques}}.
  \bibinfo{pages}{453--460}.
\newblock


\bibitem[Hertzmann et~al\mbox{.}(2001)]%
        {hertzmann2001image}
\bibfield{author}{\bibinfo{person}{Aaron Hertzmann}, \bibinfo{person}{Charles~E
  Jacobs}, \bibinfo{person}{Nuria Oliver}, \bibinfo{person}{Brian Curless},
  {and} \bibinfo{person}{David~H Salesin}.} \bibinfo{year}{2001}\natexlab{}.
\newblock \showarticletitle{Image analogies}. In
  \bibinfo{booktitle}{\emph{Proceedings of the 28th annual conference on
  Computer graphics and interactive techniques}}. \bibinfo{pages}{327--340}.
\newblock


\bibitem[H{\"o}llein et~al\mbox{.}(2021)]%
        {hollein2021stylemesh}
\bibfield{author}{\bibinfo{person}{Lukas H{\"o}llein}, \bibinfo{person}{Justin
  Johnson}, {and} \bibinfo{person}{Matthias Nie{\ss}ner}.}
  \bibinfo{year}{2021}\natexlab{}.
\newblock \showarticletitle{StyleMesh: Style Transfer for Indoor 3D Scene
  Reconstructions}.
\newblock \bibinfo{journal}{\emph{arXiv preprint arXiv:2112.01530}}
  (\bibinfo{year}{2021}).
\newblock


\bibitem[Hong et~al\mbox{.}(2022)]%
        {Hong2022Avatar}
\bibfield{author}{\bibinfo{person}{Fangzhou Hong}, \bibinfo{person}{Mingyuan
  Zhang}, \bibinfo{person}{Liang Pan}, \bibinfo{person}{Zhongang Cai},
  \bibinfo{person}{Lei Yang}, {and} \bibinfo{person}{Ziwei Liu}.}
  \bibinfo{year}{2022}\natexlab{}.
\newblock \showarticletitle{AvatarCLIP: Zero-Shot Text-Driven Generation and
  Animation of 3D Avatars}. In \bibinfo{booktitle}{\emph{Proceedings of the ACM
  SIGGRAPH}}.
\newblock


\bibitem[Huang et~al\mbox{.}(2021b)]%
        {huang2021learning}
\bibfield{author}{\bibinfo{person}{Hsin-Ping Huang}, \bibinfo{person}{Hung-Yu
  Tseng}, \bibinfo{person}{Saurabh Saini}, \bibinfo{person}{Maneesh Singh},
  {and} \bibinfo{person}{Ming-Hsuan Yang}.} \bibinfo{year}{2021}\natexlab{b}.
\newblock \showarticletitle{Learning to stylize novel views}. In
  \bibinfo{booktitle}{\emph{Proceedings of the IEEE/CVF International
  Conference on Computer Vision}}. \bibinfo{pages}{13869--13878}.
\newblock


\bibitem[Huang et~al\mbox{.}(2021a)]%
        {huang2021unsupervised}
\bibfield{author}{\bibinfo{person}{Jialu Huang}, \bibinfo{person}{Jing Liao},
  {and} \bibinfo{person}{Sam Kwong}.} \bibinfo{year}{2021}\natexlab{a}.
\newblock \showarticletitle{Unsupervised image-to-image translation via
  pre-trained stylegan2 network}.
\newblock \bibinfo{journal}{\emph{IEEE Transactions on Multimedia}}
  \bibinfo{volume}{24} (\bibinfo{year}{2021}), \bibinfo{pages}{1435--1448}.
\newblock


\bibitem[Huang and Belongie(2017)]%
        {huang2017arbitrary}
\bibfield{author}{\bibinfo{person}{Xun Huang} {and} \bibinfo{person}{Serge
  Belongie}.} \bibinfo{year}{2017}\natexlab{}.
\newblock \showarticletitle{Arbitrary style transfer in real-time with adaptive
  instance normalization}. In \bibinfo{booktitle}{\emph{Proceedings of the IEEE
  international conference on computer vision}}. \bibinfo{pages}{1501--1510}.
\newblock


\bibitem[Huang et~al\mbox{.}(2018)]%
        {huang2018multimodal}
\bibfield{author}{\bibinfo{person}{Xun Huang}, \bibinfo{person}{Ming-Yu Liu},
  \bibinfo{person}{Serge Belongie}, {and} \bibinfo{person}{Jan Kautz}.}
  \bibinfo{year}{2018}\natexlab{}.
\newblock \showarticletitle{Multimodal unsupervised image-to-image
  translation}. In \bibinfo{booktitle}{\emph{Proceedings of the European
  conference on computer vision (ECCV)}}. \bibinfo{pages}{172--189}.
\newblock


\bibitem[Huang et~al\mbox{.}(2022)]%
        {Huang22StylizedNeRF}
\bibfield{author}{\bibinfo{person}{Yi-Hua Huang}, \bibinfo{person}{Yue He},
  \bibinfo{person}{Yu-Jie Yuan}, \bibinfo{person}{Yu-Kun Lai}, {and}
  \bibinfo{person}{Lin Gao}.} \bibinfo{year}{2022}\natexlab{}.
\newblock \showarticletitle{StylizedNeRF: Consistent 3D Scene Stylization as
  Stylized NeRF via 2D-3D Mutual Learning}. In
  \bibinfo{booktitle}{\emph{Computer Vision and Pattern Recognition (CVPR)}}.
\newblock


\bibitem[Jain et~al\mbox{.}(2022)]%
        {jain2022zero}
\bibfield{author}{\bibinfo{person}{Ajay Jain}, \bibinfo{person}{Ben
  Mildenhall}, \bibinfo{person}{Jonathan~T Barron}, \bibinfo{person}{Pieter
  Abbeel}, {and} \bibinfo{person}{Ben Poole}.} \bibinfo{year}{2022}\natexlab{}.
\newblock \showarticletitle{Zero-shot text-guided object generation with dream
  fields}. In \bibinfo{booktitle}{\emph{Proceedings of the IEEE/CVF Conference
  on Computer Vision and Pattern Recognition}}. \bibinfo{pages}{867--876}.
\newblock


\bibitem[Jain et~al\mbox{.}(2021)]%
        {jain2021putting}
\bibfield{author}{\bibinfo{person}{Ajay Jain}, \bibinfo{person}{Matthew
  Tancik}, {and} \bibinfo{person}{Pieter Abbeel}.}
  \bibinfo{year}{2021}\natexlab{}.
\newblock \showarticletitle{Putting nerf on a diet: Semantically consistent
  few-shot view synthesis}. In \bibinfo{booktitle}{\emph{Proceedings of the
  IEEE/CVF International Conference on Computer Vision}}.
  \bibinfo{pages}{5885--5894}.
\newblock


\bibitem[Johnson et~al\mbox{.}(2016)]%
        {johnson2016perceptual}
\bibfield{author}{\bibinfo{person}{Justin Johnson}, \bibinfo{person}{Alexandre
  Alahi}, {and} \bibinfo{person}{Li Fei-Fei}.} \bibinfo{year}{2016}\natexlab{}.
\newblock \showarticletitle{Perceptual losses for real-time style transfer and
  super-resolution}. In \bibinfo{booktitle}{\emph{European conference on
  computer vision}}. Springer, \bibinfo{pages}{694--711}.
\newblock


\bibitem[Karras et~al\mbox{.}(2019)]%
        {karras2019style}
\bibfield{author}{\bibinfo{person}{Tero Karras}, \bibinfo{person}{Samuli
  Laine}, {and} \bibinfo{person}{Timo Aila}.} \bibinfo{year}{2019}\natexlab{}.
\newblock \showarticletitle{A style-based generator architecture for generative
  adversarial networks}. In \bibinfo{booktitle}{\emph{Proceedings of the
  IEEE/CVF conference on computer vision and pattern recognition}}.
  \bibinfo{pages}{4401--4410}.
\newblock


\bibitem[Karras et~al\mbox{.}(2020)]%
        {karras2020analyzing}
\bibfield{author}{\bibinfo{person}{Tero Karras}, \bibinfo{person}{Samuli
  Laine}, \bibinfo{person}{Miika Aittala}, \bibinfo{person}{Janne Hellsten},
  \bibinfo{person}{Jaakko Lehtinen}, {and} \bibinfo{person}{Timo Aila}.}
  \bibinfo{year}{2020}\natexlab{}.
\newblock \showarticletitle{Analyzing and improving the image quality of
  stylegan}. In \bibinfo{booktitle}{\emph{Proceedings of the IEEE/CVF
  conference on computer vision and pattern recognition}}.
  \bibinfo{pages}{8110--8119}.
\newblock


\bibitem[Kato et~al\mbox{.}(2018)]%
        {kato2018neural}
\bibfield{author}{\bibinfo{person}{Hiroharu Kato}, \bibinfo{person}{Yoshitaka
  Ushiku}, {and} \bibinfo{person}{Tatsuya Harada}.}
  \bibinfo{year}{2018}\natexlab{}.
\newblock \showarticletitle{Neural 3d mesh renderer}. In
  \bibinfo{booktitle}{\emph{Proceedings of the IEEE conference on computer
  vision and pattern recognition}}. \bibinfo{pages}{3907--3916}.
\newblock


\bibitem[Klehm et~al\mbox{.}(2014)]%
        {klehm2014property}
\bibfield{author}{\bibinfo{person}{Oliver Klehm}, \bibinfo{person}{Ivo Ihrke},
  \bibinfo{person}{Hans-Peter Seidel}, {and} \bibinfo{person}{Elmar Eisemann}.}
  \bibinfo{year}{2014}\natexlab{}.
\newblock \showarticletitle{Property and lighting manipulations for static
  volume stylization using a painting metaphor}.
\newblock \bibinfo{journal}{\emph{IEEE Transactions on Visualization and
  Computer Graphics}} \bibinfo{volume}{20}, \bibinfo{number}{7}
  (\bibinfo{year}{2014}), \bibinfo{pages}{983--995}.
\newblock


\bibitem[Kolkin et~al\mbox{.}(2019)]%
        {kolkin2019style}
\bibfield{author}{\bibinfo{person}{Nicholas Kolkin}, \bibinfo{person}{Jason
  Salavon}, {and} \bibinfo{person}{Gregory Shakhnarovich}.}
  \bibinfo{year}{2019}\natexlab{}.
\newblock \showarticletitle{Style transfer by relaxed optimal transport and
  self-similarity}. In \bibinfo{booktitle}{\emph{Proceedings of the IEEE/CVF
  Conference on Computer Vision and Pattern Recognition}}.
  \bibinfo{pages}{10051--10060}.
\newblock


\bibitem[Lee et~al\mbox{.}(2020)]%
        {lee2020drit++}
\bibfield{author}{\bibinfo{person}{Hsin-Ying Lee}, \bibinfo{person}{Hung-Yu
  Tseng}, \bibinfo{person}{Qi Mao}, \bibinfo{person}{Jia-Bin Huang},
  \bibinfo{person}{Yu-Ding Lu}, \bibinfo{person}{Maneesh Singh}, {and}
  \bibinfo{person}{Ming-Hsuan Yang}.} \bibinfo{year}{2020}\natexlab{}.
\newblock \showarticletitle{Drit++: Diverse image-to-image translation via
  disentangled representations}.
\newblock \bibinfo{journal}{\emph{International Journal of Computer Vision}}
  \bibinfo{volume}{128}, \bibinfo{number}{10} (\bibinfo{year}{2020}),
  \bibinfo{pages}{2402--2417}.
\newblock


\bibitem[Li et~al\mbox{.}(2021a)]%
        {li2021deepgcns}
\bibfield{author}{\bibinfo{person}{Guohao Li}, \bibinfo{person}{Matthias
  M{\"u}ller}, \bibinfo{person}{Guocheng Qian}, \bibinfo{person}{Itzel
  Carolina~Delgadillo Perez}, \bibinfo{person}{Abdulellah Abualshour},
  \bibinfo{person}{Ali~Kassem Thabet}, {and} \bibinfo{person}{Bernard Ghanem}.}
  \bibinfo{year}{2021}\natexlab{a}.
\newblock \showarticletitle{Deepgcns: Making gcns go as deep as cnns}.
\newblock \bibinfo{journal}{\emph{IEEE Transactions on Pattern Analysis and
  Machine Intelligence}} (\bibinfo{year}{2021}).
\newblock


\bibitem[Li et~al\mbox{.}(2017)]%
        {li2017universal}
\bibfield{author}{\bibinfo{person}{Yijun Li}, \bibinfo{person}{Chen Fang},
  \bibinfo{person}{Jimei Yang}, \bibinfo{person}{Zhaowen Wang},
  \bibinfo{person}{Xin Lu}, {and} \bibinfo{person}{Ming-Hsuan Yang}.}
  \bibinfo{year}{2017}\natexlab{}.
\newblock \showarticletitle{Universal style transfer via feature transforms}.
\newblock \bibinfo{journal}{\emph{Advances in neural information processing
  systems}}  \bibinfo{volume}{30} (\bibinfo{year}{2017}).
\newblock


\bibitem[Li et~al\mbox{.}(2021b)]%
        {li2021neural}
\bibfield{author}{\bibinfo{person}{Zhengqi Li}, \bibinfo{person}{Simon
  Niklaus}, \bibinfo{person}{Noah Snavely}, {and} \bibinfo{person}{Oliver
  Wang}.} \bibinfo{year}{2021}\natexlab{b}.
\newblock \showarticletitle{Neural scene flow fields for space-time view
  synthesis of dynamic scenes}. In \bibinfo{booktitle}{\emph{Proceedings of the
  IEEE/CVF Conference on Computer Vision and Pattern Recognition}}.
  \bibinfo{pages}{6498--6508}.
\newblock


\bibitem[Liao et~al\mbox{.}(2017)]%
        {liao2017visual}
\bibfield{author}{\bibinfo{person}{Jing Liao}, \bibinfo{person}{Yuan Yao},
  \bibinfo{person}{Lu Yuan}, \bibinfo{person}{Gang Hua}, {and}
  \bibinfo{person}{Sing~Bing Kang}.} \bibinfo{year}{2017}\natexlab{}.
\newblock \showarticletitle{Visual attribute transfer through deep image
  analogy}.
\newblock \bibinfo{journal}{\emph{arXiv preprint arXiv:1705.01088}}
  (\bibinfo{year}{2017}).
\newblock


\bibitem[Lin et~al\mbox{.}(2018)]%
        {lin2018learning}
\bibfield{author}{\bibinfo{person}{Chen-Hsuan Lin}, \bibinfo{person}{Chen
  Kong}, {and} \bibinfo{person}{Simon Lucey}.} \bibinfo{year}{2018}\natexlab{}.
\newblock \showarticletitle{Learning efficient point cloud generation for dense
  3d object reconstruction}. In \bibinfo{booktitle}{\emph{proceedings of the
  AAAI Conference on Artificial Intelligence}}, Vol.~\bibinfo{volume}{32}.
\newblock


\bibitem[Lin et~al\mbox{.}(2022)]%
        {lin2022robust}
\bibfield{author}{\bibinfo{person}{Shanchuan Lin}, \bibinfo{person}{Linjie
  Yang}, \bibinfo{person}{Imran Saleemi}, {and} \bibinfo{person}{Soumyadip
  Sengupta}.} \bibinfo{year}{2022}\natexlab{}.
\newblock \showarticletitle{Robust high-resolution video matting with temporal
  guidance}. In \bibinfo{booktitle}{\emph{Proceedings of the IEEE/CVF Winter
  Conference on Applications of Computer Vision}}. \bibinfo{pages}{238--247}.
\newblock


\bibitem[Lindell et~al\mbox{.}(2021)]%
        {lindell2021autoint}
\bibfield{author}{\bibinfo{person}{David~B Lindell}, \bibinfo{person}{Julien~NP
  Martel}, {and} \bibinfo{person}{Gordon Wetzstein}.}
  \bibinfo{year}{2021}\natexlab{}.
\newblock \showarticletitle{Autoint: Automatic integration for fast neural
  volume rendering}. In \bibinfo{booktitle}{\emph{Proceedings of the IEEE/CVF
  Conference on Computer Vision and Pattern Recognition}}.
  \bibinfo{pages}{14556--14565}.
\newblock


\bibitem[Liu et~al\mbox{.}(2018)]%
        {liu2018paparazzi}
\bibfield{author}{\bibinfo{person}{Hsueh-Ti~Derek Liu},
  \bibinfo{person}{Michael Tao}, {and} \bibinfo{person}{Alec Jacobson}.}
  \bibinfo{year}{2018}\natexlab{}.
\newblock \showarticletitle{Paparazzi: surface editing by way of multi-view
  image processing.}
\newblock \bibinfo{journal}{\emph{ACM Trans. Graph.}} \bibinfo{volume}{37},
  \bibinfo{number}{6} (\bibinfo{year}{2018}), \bibinfo{pages}{221--1}.
\newblock


\bibitem[Liu et~al\mbox{.}(2021)]%
        {liu2021editing}
\bibfield{author}{\bibinfo{person}{Steven Liu}, \bibinfo{person}{Xiuming
  Zhang}, \bibinfo{person}{Zhoutong Zhang}, \bibinfo{person}{Richard Zhang},
  \bibinfo{person}{Jun-Yan Zhu}, {and} \bibinfo{person}{Bryan Russell}.}
  \bibinfo{year}{2021}\natexlab{}.
\newblock \showarticletitle{Editing conditional radiance fields}. In
  \bibinfo{booktitle}{\emph{Proceedings of the IEEE/CVF International
  Conference on Computer Vision}}. \bibinfo{pages}{5773--5783}.
\newblock


\bibitem[Lorensen and Cline(1987)]%
        {lorensen1987marching}
\bibfield{author}{\bibinfo{person}{William~E Lorensen} {and}
  \bibinfo{person}{Harvey~E Cline}.} \bibinfo{year}{1987}\natexlab{}.
\newblock \showarticletitle{Marching cubes: A high resolution 3D surface
  construction algorithm}.
\newblock \bibinfo{journal}{\emph{ACM siggraph computer graphics}}
  \bibinfo{volume}{21}, \bibinfo{number}{4} (\bibinfo{year}{1987}),
  \bibinfo{pages}{163--169}.
\newblock


\bibitem[Ma et~al\mbox{.}(2021)]%
        {ma2021deblur}
\bibfield{author}{\bibinfo{person}{Li Ma}, \bibinfo{person}{Xiaoyu Li},
  \bibinfo{person}{Jing Liao}, \bibinfo{person}{Qi Zhang},
  \bibinfo{person}{Xuan Wang}, \bibinfo{person}{Jue Wang}, {and}
  \bibinfo{person}{Pedro~V Sander}.} \bibinfo{year}{2021}\natexlab{}.
\newblock \showarticletitle{Deblur-NeRF: Neural Radiance Fields from Blurry
  Images}.
\newblock \bibinfo{journal}{\emph{arXiv preprint arXiv:2111.14292}}
  (\bibinfo{year}{2021}).
\newblock


\bibitem[Michel et~al\mbox{.}(2021)]%
        {michel2021text2mesh}
\bibfield{author}{\bibinfo{person}{Oscar Michel}, \bibinfo{person}{Roi Bar-On},
  \bibinfo{person}{Richard Liu}, \bibinfo{person}{Sagie Benaim}, {and}
  \bibinfo{person}{Rana Hanocka}.} \bibinfo{year}{2021}\natexlab{}.
\newblock \showarticletitle{Text2Mesh: Text-Driven Neural Stylization for
  Meshes}.
\newblock \bibinfo{journal}{\emph{arXiv preprint arXiv:2112.03221}}
  (\bibinfo{year}{2021}).
\newblock


\bibitem[Mildenhall et~al\mbox{.}(2019)]%
        {mildenhall2019local}
\bibfield{author}{\bibinfo{person}{Ben Mildenhall}, \bibinfo{person}{Pratul~P
  Srinivasan}, \bibinfo{person}{Rodrigo Ortiz-Cayon},
  \bibinfo{person}{Nima~Khademi Kalantari}, \bibinfo{person}{Ravi Ramamoorthi},
  \bibinfo{person}{Ren Ng}, {and} \bibinfo{person}{Abhishek Kar}.}
  \bibinfo{year}{2019}\natexlab{}.
\newblock \showarticletitle{Local light field fusion: Practical view synthesis
  with prescriptive sampling guidelines}.
\newblock \bibinfo{journal}{\emph{ACM Transactions on Graphics (TOG)}}
  \bibinfo{volume}{38}, \bibinfo{number}{4} (\bibinfo{year}{2019}),
  \bibinfo{pages}{1--14}.
\newblock


\bibitem[Mildenhall et~al\mbox{.}(2020)]%
        {mildenhall2020nerf}
\bibfield{author}{\bibinfo{person}{Ben Mildenhall}, \bibinfo{person}{Pratul~P
  Srinivasan}, \bibinfo{person}{Matthew Tancik}, \bibinfo{person}{Jonathan~T
  Barron}, \bibinfo{person}{Ravi Ramamoorthi}, {and} \bibinfo{person}{Ren Ng}.}
  \bibinfo{year}{2020}\natexlab{}.
\newblock \showarticletitle{Nerf: Representing scenes as neural radiance fields
  for view synthesis}. In \bibinfo{booktitle}{\emph{European conference on
  computer vision}}. Springer, \bibinfo{pages}{405--421}.
\newblock


\bibitem[Mordvintsev et~al\mbox{.}(2018)]%
        {mordvintsev2018differentiable}
\bibfield{author}{\bibinfo{person}{Alexander Mordvintsev},
  \bibinfo{person}{Nicola Pezzotti}, \bibinfo{person}{Ludwig Schubert}, {and}
  \bibinfo{person}{Chris Olah}.} \bibinfo{year}{2018}\natexlab{}.
\newblock \showarticletitle{Differentiable image parameterizations}.
\newblock \bibinfo{journal}{\emph{Distill}} \bibinfo{volume}{3},
  \bibinfo{number}{7} (\bibinfo{year}{2018}), \bibinfo{pages}{e12}.
\newblock


\bibitem[Mu et~al\mbox{.}(2021)]%
        {mu20213d}
\bibfield{author}{\bibinfo{person}{Fangzhou Mu}, \bibinfo{person}{Jian Wang},
  \bibinfo{person}{Yicheng Wu}, {and} \bibinfo{person}{Yin Li}.}
  \bibinfo{year}{2021}\natexlab{}.
\newblock \showarticletitle{3D Photo Stylization: Learning to Generate Stylized
  Novel Views from a Single Image}.
\newblock \bibinfo{journal}{\emph{arXiv preprint arXiv:2112.00169}}
  (\bibinfo{year}{2021}).
\newblock


\bibitem[M{\"u}ller et~al\mbox{.}(2022)]%
        {muller2022instant}
\bibfield{author}{\bibinfo{person}{Thomas M{\"u}ller}, \bibinfo{person}{Alex
  Evans}, \bibinfo{person}{Christoph Schied}, {and} \bibinfo{person}{Alexander
  Keller}.} \bibinfo{year}{2022}\natexlab{}.
\newblock \showarticletitle{Instant Neural Graphics Primitives with a
  Multiresolution Hash Encoding}.
\newblock \bibinfo{journal}{\emph{arXiv preprint arXiv:2201.05989}}
  (\bibinfo{year}{2022}).
\newblock


\bibitem[Niemeyer et~al\mbox{.}(2021)]%
        {niemeyer2021regnerf}
\bibfield{author}{\bibinfo{person}{Michael Niemeyer},
  \bibinfo{person}{Jonathan~T Barron}, \bibinfo{person}{Ben Mildenhall},
  \bibinfo{person}{Mehdi~SM Sajjadi}, \bibinfo{person}{Andreas Geiger}, {and}
  \bibinfo{person}{Noha Radwan}.} \bibinfo{year}{2021}\natexlab{}.
\newblock \showarticletitle{RegNeRF: Regularizing Neural Radiance Fields for
  View Synthesis from Sparse Inputs}.
\newblock \bibinfo{journal}{\emph{arXiv preprint arXiv:2112.00724}}
  (\bibinfo{year}{2021}).
\newblock


\bibitem[Niemeyer and Geiger(2021)]%
        {niemeyer2021giraffe}
\bibfield{author}{\bibinfo{person}{Michael Niemeyer} {and}
  \bibinfo{person}{Andreas Geiger}.} \bibinfo{year}{2021}\natexlab{}.
\newblock \showarticletitle{Giraffe: Representing scenes as compositional
  generative neural feature fields}. In \bibinfo{booktitle}{\emph{Proceedings
  of the IEEE/CVF Conference on Computer Vision and Pattern Recognition}}.
  \bibinfo{pages}{11453--11464}.
\newblock


\bibitem[Noguchi et~al\mbox{.}(2021)]%
        {noguchi2021neural}
\bibfield{author}{\bibinfo{person}{Atsuhiro Noguchi}, \bibinfo{person}{Xiao
  Sun}, \bibinfo{person}{Stephen Lin}, {and} \bibinfo{person}{Tatsuya Harada}.}
  \bibinfo{year}{2021}\natexlab{}.
\newblock \showarticletitle{Neural articulated radiance field}. In
  \bibinfo{booktitle}{\emph{Proceedings of the IEEE/CVF International
  Conference on Computer Vision}}. \bibinfo{pages}{5762--5772}.
\newblock


\bibitem[Park et~al\mbox{.}(2021a)]%
        {park2021nerfies}
\bibfield{author}{\bibinfo{person}{Keunhong Park}, \bibinfo{person}{Utkarsh
  Sinha}, \bibinfo{person}{Jonathan~T Barron}, \bibinfo{person}{Sofien
  Bouaziz}, \bibinfo{person}{Dan~B Goldman}, \bibinfo{person}{Steven~M Seitz},
  {and} \bibinfo{person}{Ricardo Martin-Brualla}.}
  \bibinfo{year}{2021}\natexlab{a}.
\newblock \showarticletitle{Nerfies: Deformable neural radiance fields}. In
  \bibinfo{booktitle}{\emph{Proceedings of the IEEE/CVF International
  Conference on Computer Vision}}. \bibinfo{pages}{5865--5874}.
\newblock


\bibitem[Park et~al\mbox{.}(2021b)]%
        {park2021hypernerf}
\bibfield{author}{\bibinfo{person}{Keunhong Park}, \bibinfo{person}{Utkarsh
  Sinha}, \bibinfo{person}{Peter Hedman}, \bibinfo{person}{Jonathan~T Barron},
  \bibinfo{person}{Sofien Bouaziz}, \bibinfo{person}{Dan~B Goldman},
  \bibinfo{person}{Ricardo Martin-Brualla}, {and} \bibinfo{person}{Steven~M
  Seitz}.} \bibinfo{year}{2021}\natexlab{b}.
\newblock \showarticletitle{Hypernerf: A higher-dimensional representation for
  topologically varying neural radiance fields}.
\newblock \bibinfo{journal}{\emph{arXiv preprint arXiv:2106.13228}}
  (\bibinfo{year}{2021}).
\newblock


\bibitem[Park et~al\mbox{.}(2020)]%
        {park2020contrastive}
\bibfield{author}{\bibinfo{person}{Taesung Park}, \bibinfo{person}{Alexei~A
  Efros}, \bibinfo{person}{Richard Zhang}, {and} \bibinfo{person}{Jun-Yan
  Zhu}.} \bibinfo{year}{2020}\natexlab{}.
\newblock \showarticletitle{Contrastive learning for unpaired image-to-image
  translation}. In \bibinfo{booktitle}{\emph{European Conference on Computer
  Vision}}. Springer, \bibinfo{pages}{319--345}.
\newblock


\bibitem[Patashnik et~al\mbox{.}(2021)]%
        {patashnik2021styleclip}
\bibfield{author}{\bibinfo{person}{Or Patashnik}, \bibinfo{person}{Zongze Wu},
  \bibinfo{person}{Eli Shechtman}, \bibinfo{person}{Daniel Cohen-Or}, {and}
  \bibinfo{person}{Dani Lischinski}.} \bibinfo{year}{2021}\natexlab{}.
\newblock \showarticletitle{Styleclip: Text-driven manipulation of stylegan
  imagery}. In \bibinfo{booktitle}{\emph{Proceedings of the IEEE/CVF
  International Conference on Computer Vision}}. \bibinfo{pages}{2085--2094}.
\newblock


\bibitem[Peng et~al\mbox{.}(2021)]%
        {peng2021animatable}
\bibfield{author}{\bibinfo{person}{Sida Peng}, \bibinfo{person}{Junting Dong},
  \bibinfo{person}{Qianqian Wang}, \bibinfo{person}{Shangzhan Zhang},
  \bibinfo{person}{Qing Shuai}, \bibinfo{person}{Xiaowei Zhou}, {and}
  \bibinfo{person}{Hujun Bao}.} \bibinfo{year}{2021}\natexlab{}.
\newblock \showarticletitle{Animatable neural radiance fields for modeling
  dynamic human bodies}. In \bibinfo{booktitle}{\emph{Proceedings of the
  IEEE/CVF International Conference on Computer Vision}}.
  \bibinfo{pages}{14314--14323}.
\newblock


\bibitem[Pumarola et~al\mbox{.}(2021)]%
        {pumarola2021d}
\bibfield{author}{\bibinfo{person}{Albert Pumarola}, \bibinfo{person}{Enric
  Corona}, \bibinfo{person}{Gerard Pons-Moll}, {and} \bibinfo{person}{Francesc
  Moreno-Noguer}.} \bibinfo{year}{2021}\natexlab{}.
\newblock \showarticletitle{D-nerf: Neural radiance fields for dynamic scenes}.
  In \bibinfo{booktitle}{\emph{Proceedings of the IEEE/CVF Conference on
  Computer Vision and Pattern Recognition}}. \bibinfo{pages}{10318--10327}.
\newblock


\bibitem[Qi et~al\mbox{.}(2017)]%
        {qi2017pointnet}
\bibfield{author}{\bibinfo{person}{Charles~R Qi}, \bibinfo{person}{Hao Su},
  \bibinfo{person}{Kaichun Mo}, {and} \bibinfo{person}{Leonidas~J Guibas}.}
  \bibinfo{year}{2017}\natexlab{}.
\newblock \showarticletitle{Pointnet: Deep learning on point sets for 3d
  classification and segmentation}. In \bibinfo{booktitle}{\emph{Proceedings of
  the IEEE conference on computer vision and pattern recognition}}.
  \bibinfo{pages}{652--660}.
\newblock


\bibitem[Radford et~al\mbox{.}(2021)]%
        {radford2021learning}
\bibfield{author}{\bibinfo{person}{Alec Radford}, \bibinfo{person}{Jong~Wook
  Kim}, \bibinfo{person}{Chris Hallacy}, \bibinfo{person}{Aditya Ramesh},
  \bibinfo{person}{Gabriel Goh}, \bibinfo{person}{Sandhini Agarwal},
  \bibinfo{person}{Girish Sastry}, \bibinfo{person}{Amanda Askell},
  \bibinfo{person}{Pamela Mishkin}, \bibinfo{person}{Jack Clark},
  {et~al\mbox{.}}} \bibinfo{year}{2021}\natexlab{}.
\newblock \showarticletitle{Learning transferable visual models from natural
  language supervision}.
\newblock \bibinfo{journal}{\emph{arXiv preprint arXiv:2103.00020}}
  (\bibinfo{year}{2021}).
\newblock


\bibitem[Ramon et~al\mbox{.}(2021)]%
        {ramon2021h3d}
\bibfield{author}{\bibinfo{person}{Eduard Ramon}, \bibinfo{person}{Gil
  Triginer}, \bibinfo{person}{Janna Escur}, \bibinfo{person}{Albert Pumarola},
  \bibinfo{person}{Jaime Garcia}, \bibinfo{person}{Xavier Giro-i Nieto}, {and}
  \bibinfo{person}{Francesc Moreno-Noguer}.} \bibinfo{year}{2021}\natexlab{}.
\newblock \showarticletitle{H3D-Net: Few-Shot High-Fidelity 3D Head
  Reconstruction}. In \bibinfo{booktitle}{\emph{Proceedings of the IEEE/CVF
  International Conference on Computer Vision}}. \bibinfo{pages}{5620--5629}.
\newblock


\bibitem[Reiser et~al\mbox{.}(2021)]%
        {reiser2021kilonerf}
\bibfield{author}{\bibinfo{person}{Christian Reiser}, \bibinfo{person}{Songyou
  Peng}, \bibinfo{person}{Yiyi Liao}, {and} \bibinfo{person}{Andreas Geiger}.}
  \bibinfo{year}{2021}\natexlab{}.
\newblock \showarticletitle{Kilonerf: Speeding up neural radiance fields with
  thousands of tiny mlps}. In \bibinfo{booktitle}{\emph{Proceedings of the
  IEEE/CVF International Conference on Computer Vision}}.
  \bibinfo{pages}{14335--14345}.
\newblock


\bibitem[Richardson et~al\mbox{.}(2021)]%
        {richardson2021encoding}
\bibfield{author}{\bibinfo{person}{Elad Richardson}, \bibinfo{person}{Yuval
  Alaluf}, \bibinfo{person}{Or Patashnik}, \bibinfo{person}{Yotam Nitzan},
  \bibinfo{person}{Yaniv Azar}, \bibinfo{person}{Stav Shapiro}, {and}
  \bibinfo{person}{Daniel Cohen-Or}.} \bibinfo{year}{2021}\natexlab{}.
\newblock \showarticletitle{Encoding in style: a stylegan encoder for
  image-to-image translation}. In \bibinfo{booktitle}{\emph{Proceedings of the
  IEEE/CVF conference on computer vision and pattern recognition}}.
  \bibinfo{pages}{2287--2296}.
\newblock


\bibitem[Ruder et~al\mbox{.}(2016)]%
        {ruder2016artistic}
\bibfield{author}{\bibinfo{person}{Manuel Ruder}, \bibinfo{person}{Alexey
  Dosovitskiy}, {and} \bibinfo{person}{Thomas Brox}.}
  \bibinfo{year}{2016}\natexlab{}.
\newblock \showarticletitle{Artistic style transfer for videos}. In
  \bibinfo{booktitle}{\emph{German conference on pattern recognition}}.
  Springer, \bibinfo{pages}{26--36}.
\newblock


\bibitem[Schonberger and Frahm(2016)]%
        {schonberger2016structure}
\bibfield{author}{\bibinfo{person}{Johannes~L Schonberger} {and}
  \bibinfo{person}{Jan-Michael Frahm}.} \bibinfo{year}{2016}\natexlab{}.
\newblock \showarticletitle{Structure-from-motion revisited}. In
  \bibinfo{booktitle}{\emph{Proceedings of the IEEE conference on computer
  vision and pattern recognition}}. \bibinfo{pages}{4104--4113}.
\newblock


\bibitem[Schwarz et~al\mbox{.}(2020)]%
        {schwarz2020graf}
\bibfield{author}{\bibinfo{person}{Katja Schwarz}, \bibinfo{person}{Yiyi Liao},
  \bibinfo{person}{Michael Niemeyer}, {and} \bibinfo{person}{Andreas Geiger}.}
  \bibinfo{year}{2020}\natexlab{}.
\newblock \showarticletitle{Graf: Generative radiance fields for 3d-aware image
  synthesis}.
\newblock \bibinfo{journal}{\emph{Advances in Neural Information Processing
  Systems}}  \bibinfo{volume}{33} (\bibinfo{year}{2020}),
  \bibinfo{pages}{20154--20166}.
\newblock


\bibitem[Sheng et~al\mbox{.}(2018)]%
        {sheng2018deep}
\bibfield{author}{\bibinfo{person}{Bin Sheng}, \bibinfo{person}{Ping Li},
  \bibinfo{person}{Chenhao Gao}, {and} \bibinfo{person}{Kwan-Liu Ma}.}
  \bibinfo{year}{2018}\natexlab{}.
\newblock \showarticletitle{Deep neural representation guided face sketch
  synthesis}.
\newblock \bibinfo{journal}{\emph{IEEE transactions on visualization and
  computer graphics}} \bibinfo{volume}{25}, \bibinfo{number}{12}
  (\bibinfo{year}{2018}), \bibinfo{pages}{3216--3230}.
\newblock


\bibitem[Shu et~al\mbox{.}(2021)]%
        {shu2021gan}
\bibfield{author}{\bibinfo{person}{Yezhi Shu}, \bibinfo{person}{Ran Yi},
  \bibinfo{person}{Mengfei Xia}, \bibinfo{person}{Zipeng Ye},
  \bibinfo{person}{Wang Zhao}, \bibinfo{person}{Yang Chen},
  \bibinfo{person}{Yu-Kun Lai}, {and} \bibinfo{person}{Yong-Jin Liu}.}
  \bibinfo{year}{2021}\natexlab{}.
\newblock \showarticletitle{Gan-based multi-style photo cartoonization}.
\newblock \bibinfo{journal}{\emph{IEEE Transactions on Visualization and
  Computer Graphics}} (\bibinfo{year}{2021}).
\newblock


\bibitem[Srinivasan et~al\mbox{.}(2021)]%
        {srinivasan2021nerv}
\bibfield{author}{\bibinfo{person}{Pratul~P Srinivasan},
  \bibinfo{person}{Boyang Deng}, \bibinfo{person}{Xiuming Zhang},
  \bibinfo{person}{Matthew Tancik}, \bibinfo{person}{Ben Mildenhall}, {and}
  \bibinfo{person}{Jonathan~T Barron}.} \bibinfo{year}{2021}\natexlab{}.
\newblock \showarticletitle{Nerv: Neural reflectance and visibility fields for
  relighting and view synthesis}. In \bibinfo{booktitle}{\emph{Proceedings of
  the IEEE/CVF Conference on Computer Vision and Pattern Recognition}}.
  \bibinfo{pages}{7495--7504}.
\newblock


\bibitem[Tov et~al\mbox{.}(2021)]%
        {tov2021designing}
\bibfield{author}{\bibinfo{person}{Omer Tov}, \bibinfo{person}{Yuval Alaluf},
  \bibinfo{person}{Yotam Nitzan}, \bibinfo{person}{Or Patashnik}, {and}
  \bibinfo{person}{Daniel Cohen-Or}.} \bibinfo{year}{2021}\natexlab{}.
\newblock \showarticletitle{Designing an encoder for stylegan image
  manipulation}.
\newblock \bibinfo{journal}{\emph{ACM Transactions on Graphics (TOG)}}
  \bibinfo{volume}{40}, \bibinfo{number}{4} (\bibinfo{year}{2021}),
  \bibinfo{pages}{1--14}.
\newblock


\bibitem[Tretschk et~al\mbox{.}(2021)]%
        {tretschk2021non}
\bibfield{author}{\bibinfo{person}{Edgar Tretschk}, \bibinfo{person}{Ayush
  Tewari}, \bibinfo{person}{Vladislav Golyanik}, \bibinfo{person}{Michael
  Zollh{\"o}fer}, \bibinfo{person}{Christoph Lassner}, {and}
  \bibinfo{person}{Christian Theobalt}.} \bibinfo{year}{2021}\natexlab{}.
\newblock \showarticletitle{Non-rigid neural radiance fields: Reconstruction
  and novel view synthesis of a dynamic scene from monocular video}. In
  \bibinfo{booktitle}{\emph{Proceedings of the IEEE/CVF International
  Conference on Computer Vision}}. \bibinfo{pages}{12959--12970}.
\newblock


\bibitem[Wang et~al\mbox{.}(2021a)]%
        {wang2021clip}
\bibfield{author}{\bibinfo{person}{Can Wang}, \bibinfo{person}{Menglei Chai},
  \bibinfo{person}{Mingming He}, \bibinfo{person}{Dongdong Chen}, {and}
  \bibinfo{person}{Jing Liao}.} \bibinfo{year}{2021}\natexlab{a}.
\newblock \showarticletitle{CLIP-NeRF: Text-and-Image Driven Manipulation of
  Neural Radiance Fields}.
\newblock \bibinfo{journal}{\emph{arXiv preprint arXiv:2112.05139}}
  (\bibinfo{year}{2021}).
\newblock


\bibitem[Wang et~al\mbox{.}(2022)]%
        {wang2022nerfcap}
\bibfield{author}{\bibinfo{person}{Kangkan Wang}, \bibinfo{person}{Sida Peng},
  \bibinfo{person}{Xiaowei Zhou}, \bibinfo{person}{Jian Yang}, {and}
  \bibinfo{person}{Guofeng Zhang}.} \bibinfo{year}{2022}\natexlab{}.
\newblock \showarticletitle{NerfCap: Human Performance Capture With Dynamic
  Neural Radiance Fields}.
\newblock \bibinfo{journal}{\emph{IEEE Transactions on Visualization and
  Computer Graphics}} (\bibinfo{year}{2022}).
\newblock


\bibitem[Wang et~al\mbox{.}(2021b)]%
        {wang2021neus}
\bibfield{author}{\bibinfo{person}{Peng Wang}, \bibinfo{person}{Lingjie Liu},
  \bibinfo{person}{Yuan Liu}, \bibinfo{person}{Christian Theobalt},
  \bibinfo{person}{Taku Komura}, {and} \bibinfo{person}{Wenping Wang}.}
  \bibinfo{year}{2021}\natexlab{b}.
\newblock \showarticletitle{NeuS: Learning Neural Implicit Surfaces by Volume
  Rendering for Multi-view Reconstruction}.
\newblock \bibinfo{journal}{\emph{Advances in Neural Information Processing
  Systems}}  \bibinfo{volume}{34} (\bibinfo{year}{2021}),
  \bibinfo{pages}{27171--27183}.
\newblock


\bibitem[Wei et~al\mbox{.}(2021)]%
        {wei2021hairclip}
\bibfield{author}{\bibinfo{person}{Tianyi Wei}, \bibinfo{person}{Dongdong
  Chen}, \bibinfo{person}{Wenbo Zhou}, \bibinfo{person}{Jing Liao},
  \bibinfo{person}{Zhentao Tan}, \bibinfo{person}{Lu Yuan},
  \bibinfo{person}{Weiming Zhang}, {and} \bibinfo{person}{Nenghai Yu}.}
  \bibinfo{year}{2021}\natexlab{}.
\newblock \showarticletitle{Hairclip: Design your hair by text and reference
  image}.
\newblock \bibinfo{journal}{\emph{arXiv preprint arXiv:2112.05142}}
  (\bibinfo{year}{2021}).
\newblock


\bibitem[Xian et~al\mbox{.}(2021)]%
        {xian2021space}
\bibfield{author}{\bibinfo{person}{Wenqi Xian}, \bibinfo{person}{Jia-Bin
  Huang}, \bibinfo{person}{Johannes Kopf}, {and} \bibinfo{person}{Changil
  Kim}.} \bibinfo{year}{2021}\natexlab{}.
\newblock \showarticletitle{Space-time neural irradiance fields for
  free-viewpoint video}. In \bibinfo{booktitle}{\emph{Proceedings of the
  IEEE/CVF Conference on Computer Vision and Pattern Recognition}}.
  \bibinfo{pages}{9421--9431}.
\newblock


\bibitem[Yang et~al\mbox{.}(2022)]%
        {yang2022recursive}
\bibfield{author}{\bibinfo{person}{Guo-Wei Yang}, \bibinfo{person}{Wen-Yang
  Zhou}, \bibinfo{person}{Hao-Yang Peng}, \bibinfo{person}{Dun Liang},
  \bibinfo{person}{Tai-Jiang Mu}, {and} \bibinfo{person}{Shi-Min Hu}.}
  \bibinfo{year}{2022}\natexlab{}.
\newblock \showarticletitle{Recursive-NeRF: An efficient and dynamically
  growing NeRF}.
\newblock \bibinfo{journal}{\emph{IEEE Transactions on Visualization and
  Computer Graphics}} (\bibinfo{year}{2022}).
\newblock


\bibitem[Yariv et~al\mbox{.}(2021)]%
        {yariv2021volume}
\bibfield{author}{\bibinfo{person}{Lior Yariv}, \bibinfo{person}{Jiatao Gu},
  \bibinfo{person}{Yoni Kasten}, {and} \bibinfo{person}{Yaron Lipman}.}
  \bibinfo{year}{2021}\natexlab{}.
\newblock \showarticletitle{Volume rendering of neural implicit surfaces}.
\newblock \bibinfo{journal}{\emph{Advances in Neural Information Processing
  Systems}}  \bibinfo{volume}{34} (\bibinfo{year}{2021}),
  \bibinfo{pages}{4805--4815}.
\newblock


\bibitem[Ye et~al\mbox{.}(2021)]%
        {ye20213d}
\bibfield{author}{\bibinfo{person}{Zipeng Ye}, \bibinfo{person}{Mengfei Xia},
  \bibinfo{person}{Yanan Sun}, \bibinfo{person}{Ran Yi},
  \bibinfo{person}{Minjing Yu}, \bibinfo{person}{Juyong Zhang},
  \bibinfo{person}{Yu-Kun Lai}, {and} \bibinfo{person}{Yong-Jin Liu}.}
  \bibinfo{year}{2021}\natexlab{}.
\newblock \showarticletitle{3D-CariGAN: an end-to-end solution to 3D caricature
  generation from normal face photos}.
\newblock \bibinfo{journal}{\emph{IEEE Transactions on Visualization and
  Computer Graphics}} (\bibinfo{year}{2021}).
\newblock


\bibitem[Yin et~al\mbox{.}(2021)]%
        {yin20213dstylenet}
\bibfield{author}{\bibinfo{person}{Kangxue Yin}, \bibinfo{person}{Jun Gao},
  \bibinfo{person}{Maria Shugrina}, \bibinfo{person}{Sameh Khamis}, {and}
  \bibinfo{person}{Sanja Fidler}.} \bibinfo{year}{2021}\natexlab{}.
\newblock \showarticletitle{3dstylenet: Creating 3d shapes with geometric and
  texture style variations}. In \bibinfo{booktitle}{\emph{Proceedings of the
  IEEE/CVF International Conference on Computer Vision}}.
  \bibinfo{pages}{12456--12465}.
\newblock


\bibitem[Yu et~al\mbox{.}(2021a)]%
        {yu2021plenoctrees}
\bibfield{author}{\bibinfo{person}{Alex Yu}, \bibinfo{person}{Ruilong Li},
  \bibinfo{person}{Matthew Tancik}, \bibinfo{person}{Hao Li},
  \bibinfo{person}{Ren Ng}, {and} \bibinfo{person}{Angjoo Kanazawa}.}
  \bibinfo{year}{2021}\natexlab{a}.
\newblock \showarticletitle{Plenoctrees for real-time rendering of neural
  radiance fields}. In \bibinfo{booktitle}{\emph{Proceedings of the IEEE/CVF
  International Conference on Computer Vision}}. \bibinfo{pages}{5752--5761}.
\newblock


\bibitem[Yu et~al\mbox{.}(2021b)]%
        {yu2021pixelnerf}
\bibfield{author}{\bibinfo{person}{Alex Yu}, \bibinfo{person}{Vickie Ye},
  \bibinfo{person}{Matthew Tancik}, {and} \bibinfo{person}{Angjoo Kanazawa}.}
  \bibinfo{year}{2021}\natexlab{b}.
\newblock \showarticletitle{pixelnerf: Neural radiance fields from one or few
  images}. In \bibinfo{booktitle}{\emph{Proceedings of the IEEE/CVF Conference
  on Computer Vision and Pattern Recognition}}. \bibinfo{pages}{4578--4587}.
\newblock


\bibitem[Zhang et~al\mbox{.}(2022b)]%
        {zhang2022controllable}
\bibfield{author}{\bibinfo{person}{He Zhang}, \bibinfo{person}{Fan Li},
  \bibinfo{person}{Jianhui Zhao}, \bibinfo{person}{Chao Tan},
  \bibinfo{person}{Dongming Shen}, \bibinfo{person}{Yebin Liu}, {and}
  \bibinfo{person}{Tao Yu}.} \bibinfo{year}{2022}\natexlab{b}.
\newblock \showarticletitle{Controllable Free Viewpoint Video Reconstruction
  Based on Neural Radiance Fields and Motion Graphs}.
\newblock \bibinfo{journal}{\emph{IEEE Transactions on Visualization and
  Computer Graphics}} (\bibinfo{year}{2022}).
\newblock


\bibitem[Zhang et~al\mbox{.}(2022a)]%
        {zhang2022arf}
\bibfield{author}{\bibinfo{person}{Kai Zhang}, \bibinfo{person}{Nick Kolkin},
  \bibinfo{person}{Sai Bi}, \bibinfo{person}{Fujun Luan},
  \bibinfo{person}{Zexiang Xu}, \bibinfo{person}{Eli Shechtman}, {and}
  \bibinfo{person}{Noah Snavely}.} \bibinfo{year}{2022}\natexlab{a}.
\newblock \showarticletitle{ARF: Artistic Radiance Fields}.
\newblock \bibinfo{journal}{\emph{arXiv preprint arXiv:2206.06360}}
  (\bibinfo{year}{2022}).
\newblock


\bibitem[Zhang et~al\mbox{.}(2020b)]%
        {zhang2020nerf++}
\bibfield{author}{\bibinfo{person}{Kai Zhang}, \bibinfo{person}{Gernot
  Riegler}, \bibinfo{person}{Noah Snavely}, {and} \bibinfo{person}{Vladlen
  Koltun}.} \bibinfo{year}{2020}\natexlab{b}.
\newblock \showarticletitle{Nerf++: Analyzing and improving neural radiance
  fields}.
\newblock \bibinfo{journal}{\emph{arXiv preprint arXiv:2010.07492}}
  (\bibinfo{year}{2020}).
\newblock


\bibitem[Zhang et~al\mbox{.}(2020a)]%
        {zhang2020deep}
\bibfield{author}{\bibinfo{person}{Mohan Zhang}, \bibinfo{person}{Jing Liao},
  {and} \bibinfo{person}{Jinhui Yu}.} \bibinfo{year}{2020}\natexlab{a}.
\newblock \showarticletitle{Deep Exemplar-based Color Transfer for 3D Model}.
\newblock \bibinfo{journal}{\emph{IEEE Transactions on Visualization and
  Computer Graphics}} (\bibinfo{year}{2020}).
\newblock


\bibitem[Zhang et~al\mbox{.}(2021)]%
        {zhang2021nerfactor}
\bibfield{author}{\bibinfo{person}{Xiuming Zhang}, \bibinfo{person}{Pratul~P
  Srinivasan}, \bibinfo{person}{Boyang Deng}, \bibinfo{person}{Paul Debevec},
  \bibinfo{person}{William~T Freeman}, {and} \bibinfo{person}{Jonathan~T
  Barron}.} \bibinfo{year}{2021}\natexlab{}.
\newblock \showarticletitle{Nerfactor: Neural factorization of shape and
  reflectance under an unknown illumination}.
\newblock \bibinfo{journal}{\emph{ACM Transactions on Graphics (TOG)}}
  \bibinfo{volume}{40}, \bibinfo{number}{6} (\bibinfo{year}{2021}),
  \bibinfo{pages}{1--18}.
\newblock


\bibitem[Zhao et~al\mbox{.}(2014)]%
        {zhao2014parallel}
\bibfield{author}{\bibinfo{person}{Yandan Zhao}, \bibinfo{person}{Xiaogang
  Jin}, \bibinfo{person}{Yingqing Xu}, \bibinfo{person}{Hanli Zhao},
  \bibinfo{person}{Meng Ai}, {and} \bibinfo{person}{Kun Zhou}.}
  \bibinfo{year}{2014}\natexlab{}.
\newblock \showarticletitle{Parallel style-aware image cloning for artworks}.
\newblock \bibinfo{journal}{\emph{IEEE Transactions on Visualization and
  Computer Graphics}} \bibinfo{volume}{21}, \bibinfo{number}{2}
  (\bibinfo{year}{2014}), \bibinfo{pages}{229--240}.
\newblock


\bibitem[Zhu et~al\mbox{.}(2017)]%
        {zhu2017unpaired}
\bibfield{author}{\bibinfo{person}{Jun-Yan Zhu}, \bibinfo{person}{Taesung
  Park}, \bibinfo{person}{Phillip Isola}, {and} \bibinfo{person}{Alexei~A
  Efros}.} \bibinfo{year}{2017}\natexlab{}.
\newblock \showarticletitle{Unpaired image-to-image translation using
  cycle-consistent adversarial networks}. In
  \bibinfo{booktitle}{\emph{Proceedings of the IEEE international conference on
  computer vision}}. \bibinfo{pages}{2223--2232}.
\newblock


\end{thebibliography}

\end{document}